\documentclass{article}

\usepackage{arxiv}
\usepackage[utf8]{inputenc} 
\usepackage[T1]{fontenc}    
\usepackage{booktabs}       
\usepackage{amsfonts}       
\usepackage{nicefrac}       
\usepackage{microtype}      
\usepackage{lipsum}

\usepackage[numbers]{natbib}
\usepackage{graphicx, wrapfig}
\usepackage{amsmath,amssymb}
\usepackage{color}
\PassOptionsToPackage{hyphens}{url}\usepackage[hidelinks]{hyperref}
\usepackage{xcolor}
\hypersetup{
    colorlinks,
    linkcolor={red!50!black},
    citecolor={blue!50!black},
    urlcolor={blue!80!black}
}
\usepackage{caption, subcaption}
\usepackage[toc,page]{appendix}
\newcommand{\argmin}[1]{\underset{#1}{\operatorname{arg}\!\operatorname{min}}\;}

\title{A Tracking System For Baseball Game Reconstruction}
\author{
  Nina Wiedemann \\
  Department of Computer Science\\
  ETH Zürich\\
  Zürich, Switzerland \\
  \texttt{wnina@ethz.ch} \\
     \And
  Carlos Dietrich \\
  Tandon School of Engineering\\
  New York University\\
  New York, USA \\
  \texttt{cadietrich@gmail.com} \\
   \And
 Claudio T. Silva \\
  Tandon School of Engineering\\
  New York University\\
  New York, USA \\
  \texttt{csilva@nyu.edu}
}

\begin{document}
\maketitle

\begin{abstract}
The baseball game is often seen as many contests that are performed between individuals. The duel between the pitcher and the batter, for example, is considered the ``engine that drives the sport''\footnote{\url{https://ken.arneson.name/2014/11/10-things-i-believe-about-baseball-without-evidence/}}. The pitchers use a variety of strategies to gain competitive advantage against the batter, who does his best to figure out the ball trajectory and react in time for a hit.
In this work, we propose a system that captures the movements of the pitcher, the batter, and the ball in a high level of detail, and discuss several ways how this information may be processed to compute interesting statistics. 
We demonstrate on a large database of videos that our methods achieve comparable results as previous systems, while operating solely on video material.
In addition, state-of-the-art AI techniques are incorporated to  augment the amount of information that is made available for players, coaches, teams, and fans.
\end{abstract}


\begin{figure}[ht]
    \begin{subfigure}{0.5\textwidth}
        \includegraphics[width=\textwidth]{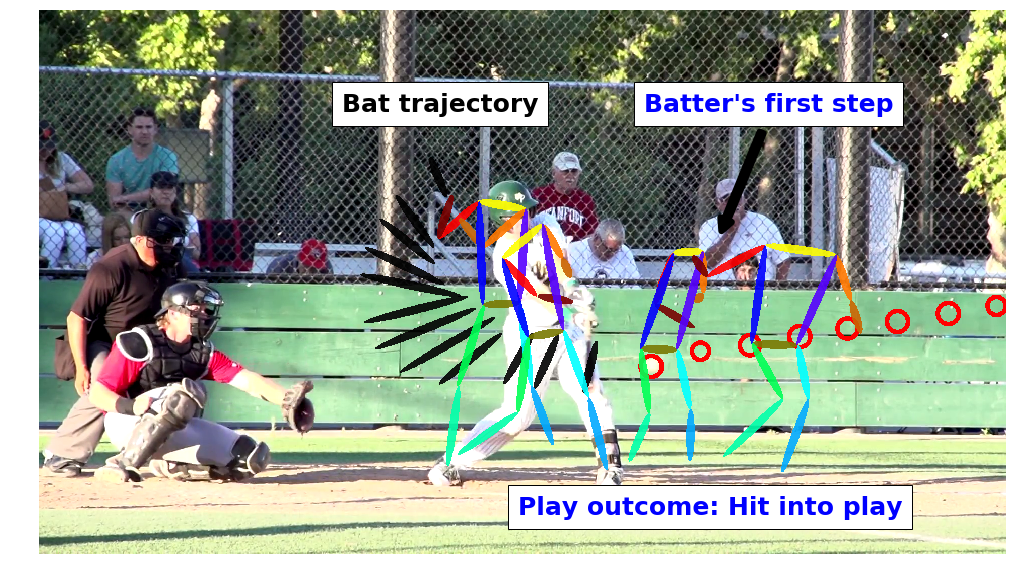}
        \caption{Batter reconstruction}
    \end{subfigure}
    \begin{subfigure}{0.5\textwidth}
        \includegraphics[width=\textwidth]{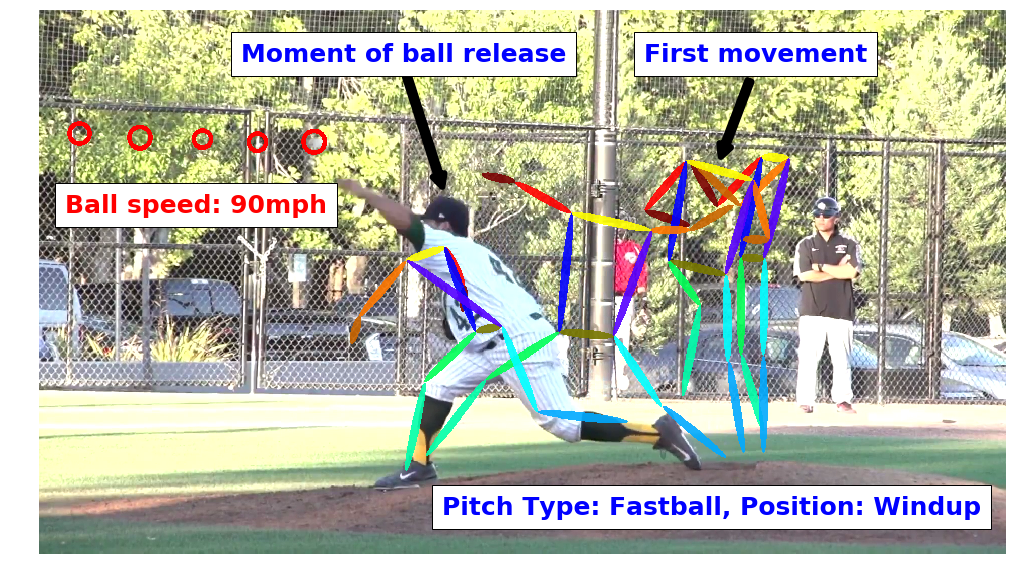}
        \caption{Pitcher reconstruction}
    \end{subfigure}
    \caption{Capturing player, ball and bat trajectories and important events from video data for game reconstruction}
    \label{fig:reconstruction}
\end{figure}

\vspace{5em}

\section{Introduction}

In recent years, the world of professional sports has been shaken by the amount of data that is being generated about players, teams and games.
Sports data is now a commodity, and everyone from fans to teams is trying to figure out the competitive advantages that can arise from properly yielding and interpreting that data. The baseball organizations, maybe more than any others, were always data-savvy. Each morning, baseball teams ``receive data bundles from the league that contains play-by-play files from the previous night’s major- and minor-league games''\footnote{\url{https://www.cbssports.com/mlb/news/the-surprising-places-mlb-teams-get-their-information-from-in-the-post-moneyball-era/}}.

The Major League Baseball (MLB) itself makes part of this data publicly available since 2008 in the Gameday server, as a collection of XML documents with data as provided by the PITCHf/x  system \cite{pitchfx}. PITCHf/x offers details about the ball trajectory for each pitch, and sets the industry standard since 2010. 
MLB Advanced Media (MLBAM), the digital arm of MLB, is the main provider of baseball data which includes information about players, the game and the teams. Since the start of the Statcast\footnote{\url{http://m.mlb.com/statcast/leaderboard\#exit-velo,p,2017}} project in 2015, MLBAM also captures the position of players, the ball and high-level game events with an unprecedented level of detail.

Although the tracking data provided by Statcast allows the analysis of the actions of the players in great level of detail, it also displays the limitations of similar tracking systems -- 
each player is only represented by a 2D coordinate over the field, and huge amounts of data (approximately 7 terabytes of uncompressed data per game) need to be transmitted and stored. The limited amount of data available from each player, associated with the cost of the installed infrastructure, results in an expensive system that is not capable of answering some of the interesting questions of the baseball community. This characteristic can be also observed on other tracking systems, usually organized around a huge infrastructure that supports the optical/radar/radiofrequency tracking of 2D player coordinates.

We believe these shortcomings can be addressed with new machine learning approaches, in particular considering recent advances in object detection, action recognition and human pose estimation. Such tools can be used to augment the amount of information that is made available for coaches, players and fans. For example, detailed player profiles comprising information about speed, reaction times and tactics can help coaches assess a player's value. On the other hand, motion models can give the players themselves more insights into their movements and help them to improve their performance. Eventually, machine learning might yield insights into the components that make a pitch or a hit successful. 

In this work, we therefore propose a new system and processing pipeline that captures the baseball game in a high level of detail. In particular, we show how the players' movements can be extracted from videos, classified, and further analyzed together with the ball trajectory to compute interesting statistics. The framework operates solely on video material, incorporating and combing state-of-the-art AI techniques to extract as much information as possible from this source. Thereby, we both improve the accuracy of statistics that have been available in previous systems, as well as extend the amount of information available. We demonstrate the relevance of the proposed system in the tracking of the actions of the most important contest in a baseball game, the pitcher-batter duel. The pitchers use a variety of pitch types (their repertoire) and tricks to gain competitive advantage against the batter, who does his best to figure out the ball trajectory and react in time for a hit. On this contest, any insight is an invaluable piece of information. As depicted in Fig. \ref{fig:reconstruction}, our system is able to capture and reconstruct the interaction between pitcher and batter extensively. We provide detailed information about their body movements as well as high-level descriptions of strategies and game events. 

The main steps of our system may be summarized as:
\begin{description}{}

\item[1.] The tracking of the stance of the players, where the stance is represented by eighteen body key points (or joints) for each frame;
\item[2.] The processing of the joint trajectories to classify player actions;
\item[3.]  The tracking of the ball and bat;
\item[4.] The processing of both joint trajectories and the detection of fast moving objects to find key events in the game (e.g. start of the play).

\end{description}

We consider as our main contribution the design of a fully self-contained system for baseball game reconstruction. Here we present the framework in the following steps: First, in chapter \ref{terminology} we provide relevant background information regarding baseball analysis, previously implemented systems and related work on pose estimation and object detection. We then provide an overview of the system in chapter \ref{overview}, where we shortly describe each module, while a detailed explanation of each single method can be found in appendix \ref{joint_tracking}-\ref{object_detecion_methods}. Furthermore, in chapter \ref{results} the performance of all parts of the framework is evaluated on data capturing the battle between pitcher and batter. After the integration of units in the larger system is outlined in chapter \ref{integration_legotracker} together with details on implementation and performance, we discuss the current results and possible directions of further research in chapter \ref{discussion}. 

\section{Background} \label{terminology}

\subsection{Problem statement}

Game reconstruction requires to extract as much information as possible from videos. The more detailed the movement of players is recorded, the more analysts can conclude about the success of certain motion patterns and tactics. On the other hand, play diagrams are constructed representing when and where an important event happened in the game \cite{ono2018baseball}. Coaches and fans use such tools to evaluate game mechanics. Thus, various systems have been installed that aim to capture the data that later serves as input to the statistical analysis. Sports data tracking systems usually rely on a subset of three different kinds of input: optical (video data), wearable sensors and radar (or depth cameras). A classic example of an optical motion tracking system is PITCHf/x \cite{pitchfx}, which computes the ball trajectory from three video cameras on Major League Baseball (MLB) venues. The information about ball spin and speed provided by PITCHf/x can already be used to cluster and classify pitch types, as shown by \citet{Pane2013}.

\paragraph{Statcast}
However, the PITCHf/x system was replaced with the developement of the Statcast system \footnote{\url{http://m.mlb.com/statcast/leaderboard\#exit-velo,p,2017}}. Since 2015 it is installed in all major league venues and captures significantly more information than before. With a combination of optical cameras and speed radars, all players on the field and the ball are tracked. The goal of the radar is to enable more precise description of the ball trajectory, including spin rate and velocity, and to gain information about player speed and interaction with the ball (e.g. the moment of ball release). The system also uses manual input by operators to tag extra events and assess data consistency. The output of the StatCast system, i.e. discrete player and ball positions over time, can be used for game reconstruction to visually explore plays, for example in the framework of \citet{dietrich14}.

Even though Statcast improved game reconstruction a lot, we believe that it is still possible to gain more insights. Most importantly, the players are only tracked as points on the field, not providing any information about the movement of single body parts. Tracking itself is hard, and in other sports it is usually based on inertial sensors such as GPS and RFID tags \cite{Winter2016, Lee2012, Ride2012, Wisbey2010}. For American football, \citet{Foing2010} describe a system that analyzes RFID sensor data input to yield player analyses in three stages. But even if they started equipping all players with a sensor in baseball as well, no detailed information about the body motion of a single player during pitch or swing would be gained. This would require sensors in arms and legs.

\subsection{Baseball game terminology}

Before discussing related work from computer vision that can help to fill the gap, some terminology of baseball must be established. Since we focus on the interaction between the pitcher and the batter, only related terminology will be explained here. For a detailed explanation of baseball rules, please refer to \citet{baseballrules}. A baseball game is divided into nine \textit{innings}, which all consist of a certain number of \textit{plays} (called at-bats). A \textit{play} starts with the \textit{pitcher} of team A throwing (\textit{pitching}) the ball from his position at the \textit{pitcher's mound} to the \textit{batter} of team B. If the ball passes through a defined area between the batters knees and hips, it is either a \textit{strike} or the batter can \textit{swing} the \textit{bat} and the ball is \textit{hit into play}, which means that all \textit{runners} of team B are allowed to move from \textit{base} to base, until they reach \textit{home plate}. They need to reach a base before the pitcher's team has passed the ball to a certain position.

\subsection{Motion analysis}\label{motion_analysis}

In the process of a play as described above, a detailed motion analysis can be beneficial on several parts: Firstly, during the pitch itself, different \textit{pitching positions} and \textit{pitch types} can be distinguished. Regarding the delivery, pitchers (starters) usually start facing the plate, and make a sideways step to get into proper a position (\textit{Windup}). When there are runners on bases, they pitch from the \textit{Stretch}, which is a technique to shorten the motion and give the runners less time to steal bases. Detecting runners is not sufficient to determine the pitching position though, as some pitchers vary the position unexpectedly. For analysts trying to grasp the behaviour of a player it is therefore crucial to know how often and when the player did a Windup/a Stretch. Furthermore, pitchers can throw the ball in different ways to make it harder for the batter to hit. These \textit{pitch types} differ in the pitcher's hand grip when holding the ball, affecting ball velocity and spin. Similar to the position, a comprehensive tracking system should infer this information directly from the video data.

Secondly, not only the classification of specific movement (e.g. pitch type classification) is relevant for analysis, but also the overall body motion. In the long run it might be possible to construct a 3D model of the motion, which makes the motion of different players comparable and might yield insights into what makes a player successful. Thus, we aim to break down the representation of a player's motion into just a view coordinates describing the displacement of important body parts. More specifically, we believe that computer vision methods for pose estimation must be incorporated in a modern system for baseball analysis.

\paragraph{Pose estimation}

Human pose estimation involves finding a set of coordinates depicting the location of the body parts (or joints) of each person in the image. Performing such inference on each frame of the baseball videos, a player’s motion is described as the trajectories of their body parts over time. Instead of the 2D position of the player on the field, we get a low dimensional description of the displacement of certain body key points comprising legs, arms, hips and shoulders.

The large majority of frameworks simply 
Frameworks can be grouped using video as input or single images, and operating in top down (firstly a person detector is used, then pose estimation applied), or bottom up (joints coordinates are detected and then assigned to people) fashion to handle multiple people. According to the COCO (Common Objects in Context) keypoint challenge 2017, a CNN by \citet{chen} is state-of-the-art in human pose estimation in 2D images, followed by similar approaches (\cite{fang2017}, \cite{Papandreou} and \cite{pose_estimation}). \citet{RNN_pose} on the other hand use a Recurrent Neural Network (RNN). 

The use of videos instead of single images may help to reduce the uncertainty on the pose estimation, since the temporal consistency provides additional information for the detection of keypoints. \citet{chained_pose} employ a recurrent connection between separate CNNs operating on single images. Usually though in the literature, a different kind of RNN, so-called Long-Short-Term-Memory (LSTM) cells, are used for sequential data. Regarding pose estimation, for example \citet{lstm_pose} directly use an LSTM on video input.

Recently also 3D pose estimation has been developed, and different methods were discussed by \citet{3Devaluation}, including \cite{Yasin2016,Li2015,Li2014, Kostrikov2014}. Using CNNs, \citet{vnect} achieve remarkable results with a model trained on a data set from the Max-Planck-Institute \cite{MPI}. The model is only applicable for a single person though.

In the first version of our proposed framework we employ the model by \citet{pose_estimation}, because for a long time it was by far the fastest one, performing 2D pose estimation at a rate of 8.8 frames per second (fps) for images of size 368-by-654. Furthermore, running time is independent of the number of people in the image, which is important for sports videos with audience in the background. Note however that in the modular fashion in which the system is described, the approach can be replaced by new state of the arts methods at anytime.

\paragraph{Action recognition}

While pose estimation is valuable on its own for game reconstruction, another goal is to classify motion based on the pose data, for example to distinguish pitch types as explained above. Most previous work on action recognition such as \cite{Masurelle} or \cite{Wang} require depth cameras or other aids though. On the other hand, lots of work is available classifying actions from videos, but as videos require larger computational effort, it is preferable to work on the processed pose data instead. To our knowledge, only in \cite{basketball} such an approach is implemented, i.e. machine learning techniques are applied on the 2D pose estimation keypoints of the players. Here, they predict whether a throw in basketball resulted in a miss or score.

\subsection{Object detection}\label{object_detection_background}

While the Statcast system is able to estimate ball speed with high precision, we aim to avoid the large infrastructure of radar systems and instead incorporate object detection methods for video input for tracking ball and bat. In addition, a new piece of information that will be valuable for analysts is the position of the glove of the catcher, and more importantly, the movement of the bat during the swing.

Fortunately, baseball bats and gloves are included in the popular COCO data set, which is often used to train Artificial Neural Networks (ANNs), so work on detecting these baseball related objects is available. For example, \citet{Ren} follow up previous work (\cite{Girshick2014,Girshick2015}) to improve a successful approach called Faster R-CNN. Objects are located and labeled by first extracting regions and then predicting the probability of appearance for the object in this region. Another dataset with baseball bats called HICO-DET is used by \citet{Chao}, who train CNNs to recognize human object interactions including baseball bats. 

However, our experiments showed that these object detection methods are not able to detect blurred balls and bats during the swing. Thus, conventional approaches for object tracking are required additionally. Approaches for motion detection are compared in \cite{Wu}, evaluating work by \citet{zhong,Hare2011} and \citet{Kalal2012} as the most successful ones. However, most of their data did not include images with objects of high velocity, which usually appear in single images only as blurred semi transparent streaks. Therefore, recently \citet{Rozumnyi2017} have developed a new data set, calling these kind of images ''fast moving objects'' (FMO). To track FMOs, the authors propose a three stage system based on difference images. Since baseball is very fast, this approach is used as a building block.

\subsection{Game events}

\begin{figure}[t]
\begin{subfigure}{0.15\textwidth}
\includegraphics[width=\textwidth]{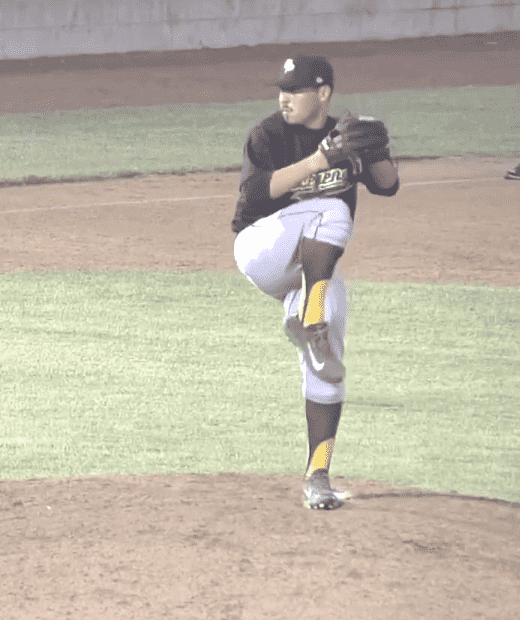}
\caption{\label{fig:events1}}
\label{fig:pitcher_raise_leg}
\end{subfigure}
\hfill
\begin{subfigure}{0.15\textwidth}
\includegraphics[width=\textwidth]{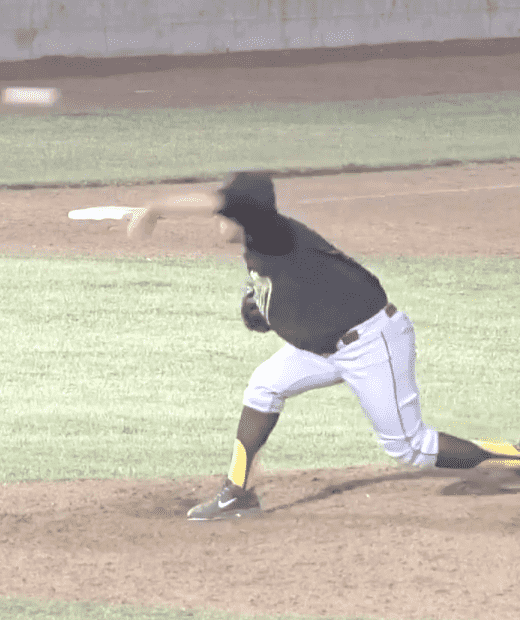}
\caption{\label{fig:events2}} 
\label{fig:pitcher_release}
\end{subfigure}
\hfill
\begin{subfigure}{0.15\textwidth}
\includegraphics[width=\textwidth]{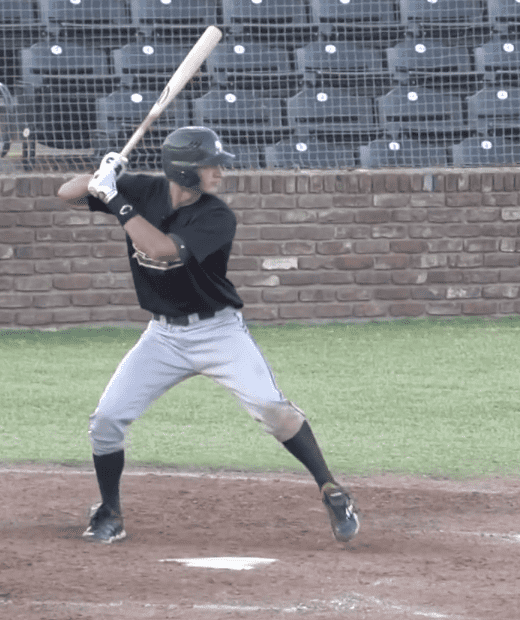}
\caption{\label{fig:events3}} 
\label{fig:batter_raise_foot}
\end{subfigure}
\hfill
\begin{subfigure}{0.15\textwidth}
\includegraphics[width=\textwidth]{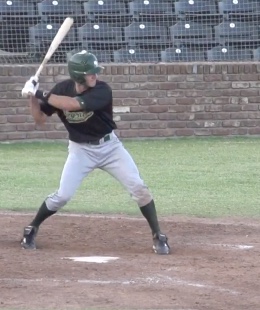}
\caption{\label{fig:events4}} 
\label{fig:batter_foot_down}
\end{subfigure}
\hfill
\begin{subfigure}{0.15\textwidth}
\includegraphics[width=\textwidth]{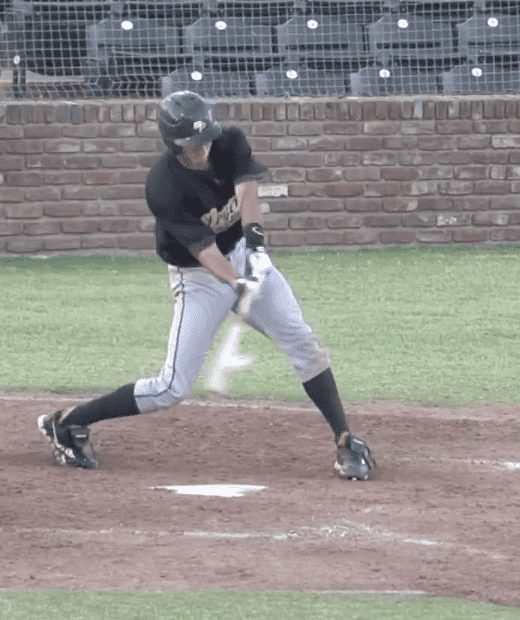}
\caption{\label{fig:events5}} 
\label{fig:batter_impact}
\end{subfigure}
\hfill
\begin{subfigure}{0.15\textwidth}
\includegraphics[width=\textwidth]{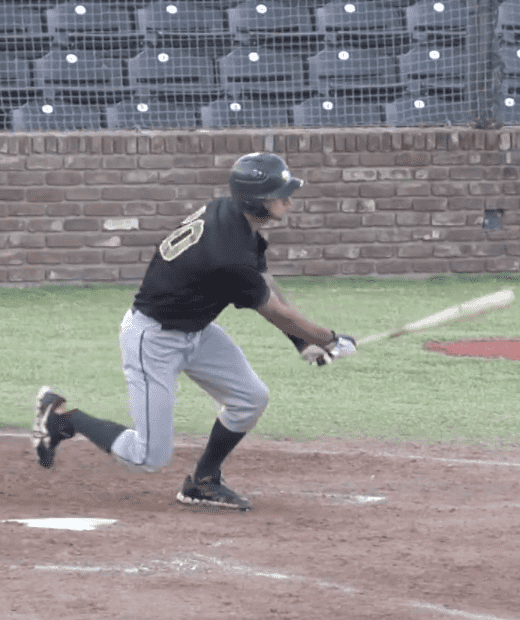}
\caption{\label{fig:events6}} 
\label{fig:batter_starts_run}
\end{subfigure}
\caption{Interesting events during one play: (\subref{fig:events1}) Pitcher raises his leg (first move), puts the leg back on the ground and releases the ball (\subref{fig:events2}). Meanwhile, the batter lifts his foot slightly (\subref{fig:events3}), and starts swinging (\subref{fig:events4}). If the ball is hit (\subref{fig:events5}) into play, the batter starts to run (\subref{fig:events6}).} \label{fig:events}
\end{figure}

Finally, the detection of additional game events may help the analysis of the \textit{play}, especially the ones that give us more information about the actions of the pitcher and the batter. We start with determining the moment the play starts, which is called \textit{pitcher's first movement} here. Since the first movement is not well-defined, we focus on finding the moment the pitcher raises his leg (Fig.~\ref{fig:events1}). Further on, the important part of the pitcher's motion ends with the \textit{ball release} (Fig.~\ref{fig:events2}). From then on we track the ball and estimate its speed.

On the plate, the batter starts to move slightly before \textit{ball release}, when he shortly lifts his foot and starts the swing (Fig.~\ref{fig:events3}--Fig.~\ref{fig:events5}). The movement of the batter may even give us hits about the \textit{play outcome}, i.e. whether the ball was \textit{hit into play}. Last, the moment the batter starts to run (his first step towards 1st base, shown in Fig.~\ref{fig:events6}) can be assessed for reaction time purposes. 

To summarize, game reconstruction involves information of several domains, including human pose estimation, action recognition and object tracking. For each part, we extract and process information from video input alone, employing several computer vision methods taken and adapted from the research that was mentioned above.

\section{Framework for baseball game reconstruction} \label{overview}

We propose a system that is similar to Statcast in scope, but based on video input alone. Additionally, we aim to automatize tasks that have previously required user input, and provide new information that has not yet been available. While in this contribution we mainly focus on describing the analysis framework necessary to realize this goal, it is important to understand the setup that we assume our software will be running on. Specifically, the hardware is planned out as a system of distributed blocks, thus called the ''Legotracker''. The lego blocks, each containing a camera and a processing unit, are spread across the field, in order to acquire high quality videos from different viewpoints. On the blocks, parts of the data processing pipeline can be executed locally. For example, if several blocks run a computer vision algorithm to register an event (e.g. time point of ball release), the information is valuable for synchronization purposes.  In general, blocks communicate with each other through a monitoring system, and send their processed data to a shared database, where more time-consuming analysis can be executed later. This way, the system requires less infrastructure and provides more detailed information about individual players due to the proximity of one of the distributed cameras.

\begin{figure}[ht]
    \begin{subfigure}{0.55\textwidth}
        \centering
        \includegraphics[width=\textwidth]{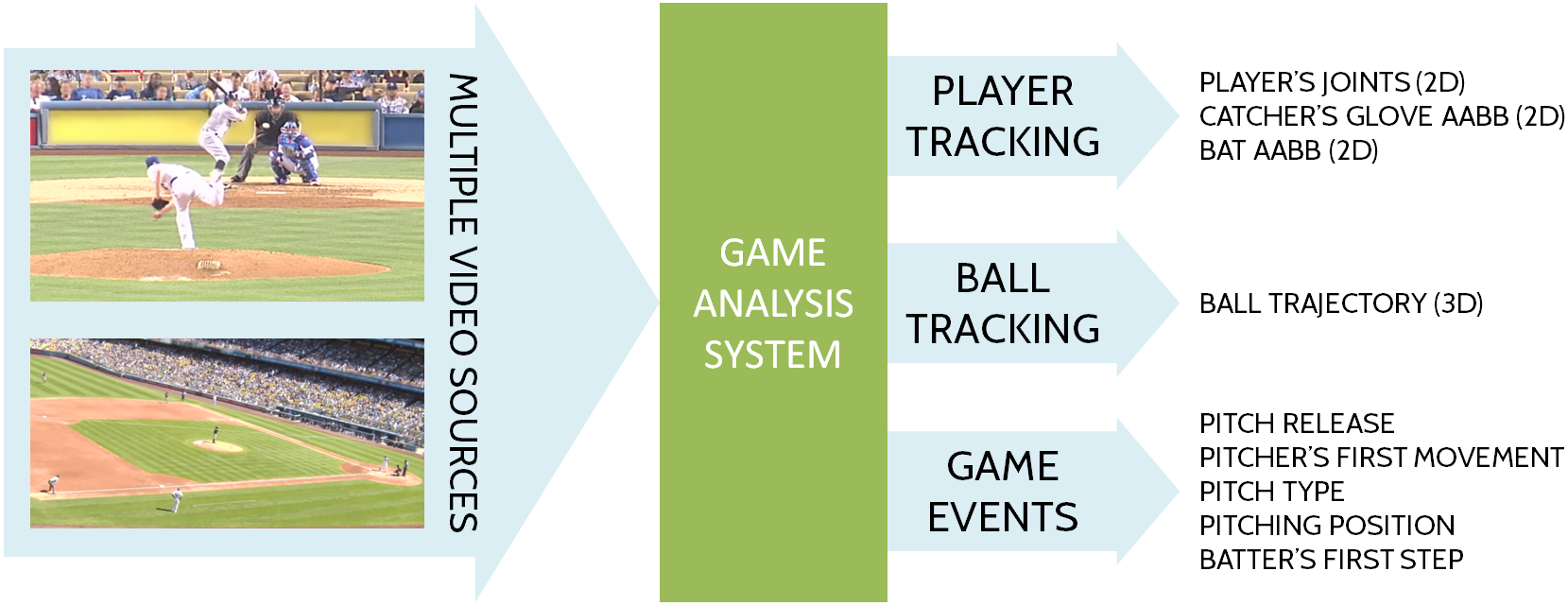}
        \caption{Video input is processed to yield three kinds of output: Player movement, ball and bat trajectories, and the time point of important game events.}
        \label{fig:system_overview}
    \end{subfigure}
    \hspace{0.03\textwidth}
    \begin{subfigure}{0.42\textwidth}
        \centering
        \includegraphics[width=\textwidth]{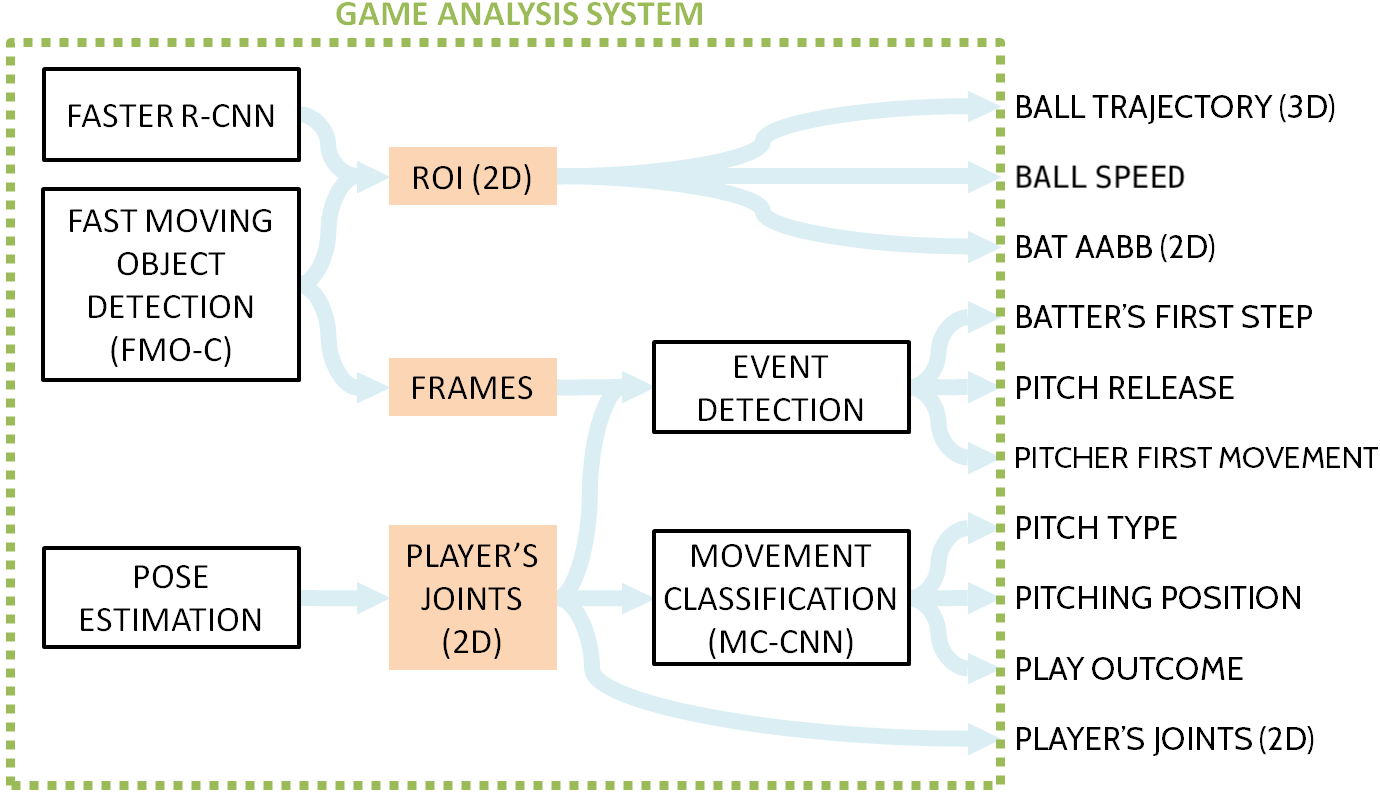}
        \caption{State-of-the-art method in pose estimation and object detection are combined, adapted and extended to provide all information necessary for baseball game reconstruction.}
        \label{fig:game_analysis_system}
    \end{subfigure}
    \caption{Overview of the processing pipeline}
    \label{fig:processing_units}
\end{figure}

The goal of the overall system is to extract (1) movement of players, (2) time points of events and (3) information about the ball and bat trajectory from video sources (Fig. \ref{fig:system_overview}). The components of the proposed framework are shown in Fig. \ref{fig:processing_units}, which is explained from bottom to top in the following. For more details on methods refer to the corresponding part of the appendix indicated for each component.

As a main component of the system, we incorporate a pre-trained model for multi-person real-time pose estimation proposed by \citet{pose_estimation}. It yields the coordinates of body joints of all persons in the frame (cf. appendix \ref{roi}). The resulting time series data can further serve as input for movement classification models as well as for event detection (e.g. to determine when the pitcher raises his leg to initiate the pitch). First, however, the target person must be distinguished from other persons in the frame, and the trajectories are imputed and smoothed with low-pass filtering (cf. appendix \ref{localization}). On the clean data, i.e. imputed single-person trajectories, deep learning techniques can be applied to classify the movement into certain categories. In our implementation this module is a 1D-CNN that we call MC-CNN. Whilst the network can be trained generically to classify any body joint trajectories of any player, it is demonstrated on three important tasks here: Regarding the pitcher's motion, MC-CNN is trained to predict pitch type and pitching position, while the batter's trajectories are used to determine the play outcome, i.e. whether the batter starts to run. For details on methodology and model architecture see appendix \ref{mccnn}.

For the other two main components of the system (event detection and object detection) we developed a difference image based approach for fast moving object (FMO) detection inspired by \citet{Rozumnyi2017}. The proposed method FMO-C thresholds difference images to output a set of ''candidates'' in each frame, indicating areas where motion occurred (cf. appendix \ref{fmoc}). 

All approaches, FMO-C, pose estimation or both combined yield the timing of the game events. This comes from the fact that some events can be described by the displacement of certain body parts and/or the motion of an object. Firstly, the pitcher's first movement can be viewed as the first series of consecutive frames in which motion is detected at the pitcher's leg (cf. appendix \ref{first_move}). Similarly, the batter's first step is detected as a significant increase of motion close to his legs. Last, the time point of pitch release (when the ball leaves the pitcher's hand) can in principle be determined analyzing when the pitcher's arm is moving the fastest. However, due to the poor quality of the available data, the arm is too blurry during the pitch, so pose estimation often fails. A more reliable way is therefore tracking the ball itself and thereby inferring when it must have been released. 

This leads to the last module of the framework dealing with ball and bat tracking. The ball is detected as a certain pattern of motion candidates which are the output of the FMO-C method. A metric is constructed that decides when the candidates in three consecutive frames are likely to correspond to a ball trajectory. In particular, it can be assumed that the trajectory is rather a straight line and the size of the ball does not vary much. For details on computation and parameters see appendix \ref{gbcv}.
Of course, so far the pipeline only yields the 2D trajectory on images. To reconstruct the 3D trajectory and estimate speed, the outputs of two synchronized cameras are compared.

Finally, FMO-C is complemented by an object detection model for bat tracking. Again we incorporate state-of-the-art methods, namely a two stage CNN for object detection called Faster R-CNN \cite{Ren}. Since baseball bats are included in the COCO dataset, a pre-trained model reliably detects the bat when it is not in motion. During the swing though, the bat becomes very blurry as well, and only FMO-C can detect it. The true motion candidate is found by comparing the candidates' location to the bat position in the previous frame, or to the last detection of Faster R-CNN when it has just started moving (cf. appendix \ref{bat}).

Our main contribution is the construction of the overall framework and processing pipeline. While all single components consist of adapted and combined previous work, the work here describes how they are plugged together to yield a system suitable for the specific situation in baseball, considering the available data and the large variety of analysis tasks. 

\section{Results}\label{results}

\subsection{Data for training and testing}

\begin{figure}[ht]

\begin{subfigure}{0.5\textwidth}
    \centering
    \includegraphics[height=3cm]{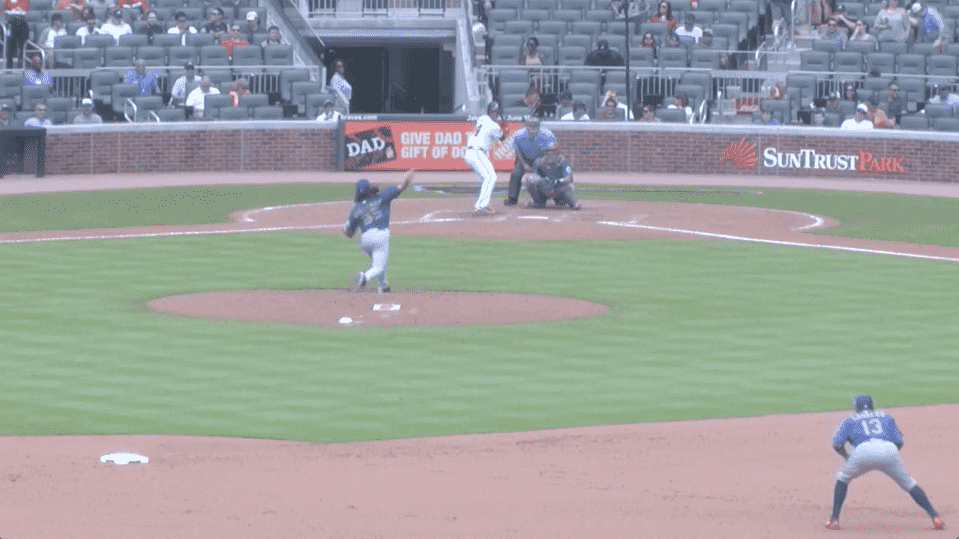}
    \caption{center-field camera} 
    \label{fig:center_field}
\end{subfigure}
\hfill
\begin{subfigure}{0.5\textwidth}
    \centering
    \includegraphics[height=3cm]{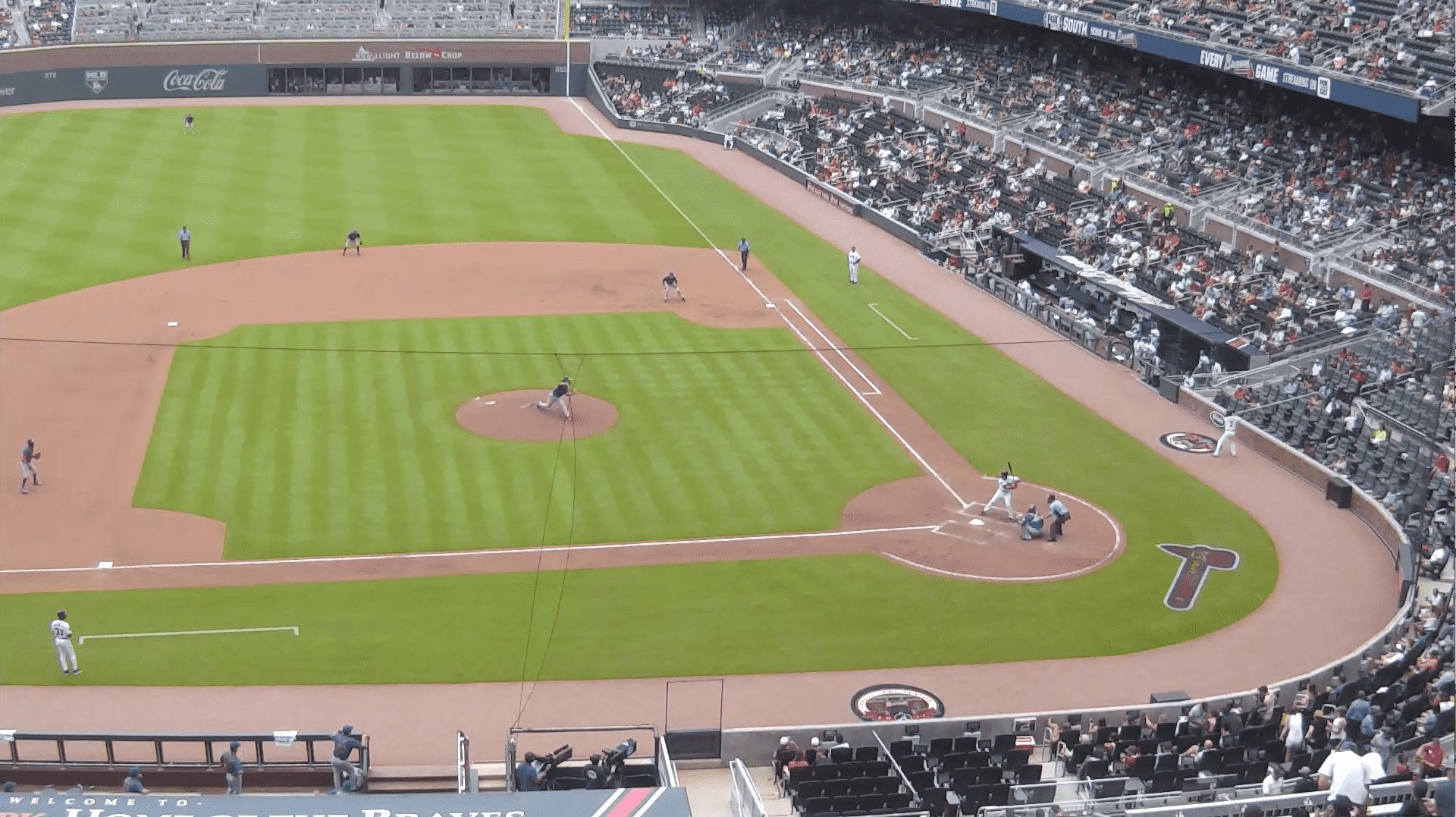}
    \caption{side-view camera} 
    \label{fig:side_view}
\end{subfigure}
\caption[Videos of the MLBAM auditing tool (data available through a collaboration with NYU)]{Input videos are non-synchronized videos from two different viewpoints}
\label{cameraviewpoints}
\end{figure}

Since the Legotracker hardware is not used in practice so far, the dataset used here is quite different from the data we plan to acquire in the end. We consider the tests on this data as a stress test for our methods, because the Legotracker data will probably be of better quality due to closer cameras, more available viewpoints per person and better temporal resolution. Since the Statcast system is based on radar, of course the video quality is not as important. 

We tested the system with data captured from two viewpoints: one in the center-field (Fig.~\ref{fig:center_field}), focusing the home plate, and one in a high-side view of the infield (Fig.~\ref{fig:side_view}). However, it is not trivial to combine both views as the cameras are not synchronized. One video does not comprise a whole game, but corresponds only to one play, including all action between pitcher and batter. The center field videos are cut to 6.5 seconds length (around 165 frames, as the frame rate is always 30fps), roughly aligned as the ball release occurs always around frame 90. The side view videos are less aligned and often longer (up to 300 frames). In general though the start of the action and the time points of other events vary widely. Showing games with more than 200 different pitchers and more than 300 batters, the dataset is very diverse. In addition to videos, metadata was available providing the initial position of the target players, the pitch type, pitching position, play outcome, ball release frame index and the pitcher's first movement frame index.

\subsection{Joint tracking}\label{mc}

In the domain of baseball analysis, it is important to note the difference between joint detection and joint tracking. While the system requires real-time continuous output, most frameworks including OpenPose \cite{pose_estimation} do inference frame by frame. This leads to two main issues: Firstly, the output coordinates do not transition smoothly, but sometimes jump between two consecutive frames. Secondly, a person is not identified by appearance, so the coordinates of a person in one frame are not associated with the position in the previous frame. Thus, we first describe the experiments conducted on joint tracking, involving joint detection, player localization and filtering of the trajectories.

\paragraph{Pose estimation}

\begin{figure}[t]
    \centering
    \begin{subfigure}{0.31\textwidth}
        \includegraphics[width=\textwidth]{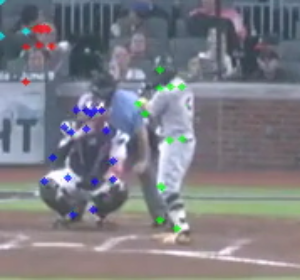}
        \caption{\label{fig:localize_a}}
    \end{subfigure}
    \hfill
    \begin{subfigure}{0.31\textwidth}
        \includegraphics[width=\textwidth]{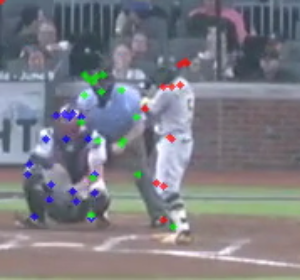}
        \caption{\label{fig:localize_b}}
    \end{subfigure}
    \hfill
    \begin{subfigure}{0.31\textwidth}
        \includegraphics[width=\textwidth]{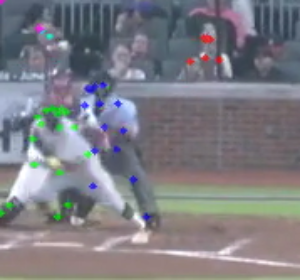}
        \caption{\label{fig:localize_c}}
    \end{subfigure}
    \caption{Challenges in tracking the target player: The output of the pose estimation model can be quite different even in consecutive frames. For example, from (\subref{fig:localize_a}) to (\subref{fig:localize_b}) the order of detected people changes (person index in output list corresponds to colour) and a new person (the referee) is detected correctly. Furthermore, the output is not very accurate on blurry frames, and people are sometimes mixed up as shown in (\subref{fig:localize_c}).}
    \label{fig:localization_problems}
\end{figure}

As already mentioned, for frame-by-frame detection of body parts we apply a pre-trained model for pose estimation \cite{pose_estimation}, yielding 2D coordinates of 18 joints for all detected individuals (Fig. \ref{fig:localization_problems}). For real time performance, the images are down sampled to 368-by-654. In first tests, we observed strong artifacts, which are caused by upsampling of the pose estimation outputs. Consequently it was preferable to feed only the important part of the image to the pose estimation network, thereby making the input smaller in general, leading to less down- and upsampling. To achieve this goal, a dynamic region of interest (ROI) around the target player is computed (for details see \ref{roi}). Only very small artifacts remain and completely disappear with low-pass filtering.

Unfortunately there is no ground truth data for pose estimation in the available baseball data, impeding a quantitative assessment of the applied methods. From manual observation it can be concluded though that the model generalizes very well, even to the new domain of sports players and positions. Only in extreme poses, for example when the pitcher raises his leg very high, the network fails to associate the body parts correctly. On blurry images as in Fig. \ref{fig:localization_problems} the output is not reliable and can fail to distinguish overlapping people. On single frames this might even be hard for a human observer though.

As a more quantitative performance measure one can compute the ratio of missing values. In the output, the set of coordinates is zero if a joint could not be detected. This occurs most often for facial key points, wrists and elbows: In more than 60\% of the frames, eyes or nose are missing. The wrist can not be detected in 28\% of the frames and elbows in around 10\% of the frames on average. While for our purposes facial key points can be discarded, elbow and wrists are important for swing and pitch analysis. Problematically, these gaps occur mostly in crucial moments, for example during ball release because the arm moves so fast that it appears blurry on the frame. These problems indicate that for the final Legotracker system, it might be necessary to replace the pose estimation module with a different approach, for example one that is also using temporal information instead of single frames. On the other hand, cameras with better temporal resolution \cite{gallego2019event}, adapted exposure time and closer viewpoint might be sufficient to handle such problems as well.

\paragraph{Player localization}\label{localization_results}

Although only a ROI around the target player is fed to the model, there might still be multiple people detected, for example the referee and catcher standing directly behind the batter. Also, the number and order of detected people in the pose estimation output is not constant (compare colours in Fig. \ref{fig:localize_a} and Fig. \ref{fig:localize_b}), and overlapping people are sometimes mixed up (Fig. \ref{fig:localize_c}). We propose to take the intersection over union of bounding boxes around the joints of a player, because the results were more stable than for example comparing the distances of joint coordinates directly. The full processing procedure is described in section \ref{localization}.

For the pitcher, the approach works for all videos, so if the pose estimation network picked up people in the audience, the algorithm correctly decided to use the pitcher's joints instead of their joints. Regarding the batter, in approximately 10\% of the videos in which he starts to run (after a successful hit), at some point he is confused with the referee standing behind him. Analyzing these results we inferred that pose estimation is too inaccurate to track the batter correctly in these situations. As shown in Fig.~\ref{fig:localize_c}, pose estimation returns a set of coordinates for one person, where a few joints belong to the batter and a few to the referee (green dots). Often, the batter is not detected/separated from other people for up to 20 consecutive frames. As a result, once the batter is detected correctly again, the tracking procedure cannot determine correctly which detected person corresponds to the target person anymore. The problem is thus rather due to the pose estimation than a problem of the localization algorithm. A straightforward solution would for example be applying a more reliable person detection algorithm on top, that does not struggle from problems relates to single joint detection.

\paragraph{Interpolation and filtering}\label{filtering_results}

In the current state of the framework, missing values in the pose estimation output are removed with simple linear interpolation since other methods are unstable when larger gaps occur in the sequences. Even deep learning techniques to predict the next value in the sequence were explored, but while the prediction for one frame is very accurate, performance decreases significantly when joints are not detected for several frames which is quite common.

\begin{figure}[ht]
\begin{subfigure}{0.5\textwidth}
\includegraphics[width=\textwidth]{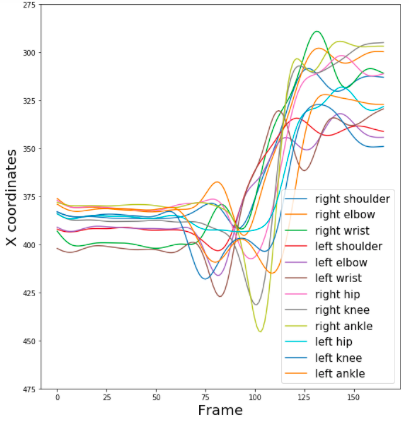}
\caption{Butter lowpass filtering}
\label{fig:lowpass}
\end{subfigure}
\begin{subfigure}{0.5\textwidth}
\includegraphics[width=\textwidth]{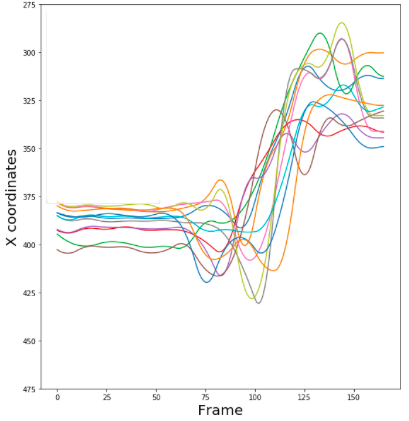}
\caption{B-spline fitting}
\label{fig:bspline}
\end{subfigure}
\caption[Self-created figure]{The output of the full pose-tracking processing pipeline is pictured, which is a time series of coordinates for each joint of one player (in this example the batter).  Here, for the sake of simplicity only the X coordinate is plotted. One can see that the whole body is moving to the left (decreasing X coordinate for all joints). The strong motion of arms and legs right before that correspond to the swing. One can compare the output of different signal filtering methods, namely Butter lowpass filtering (\subref{fig:lowpass}) and B-spline curve fitting (\subref{fig:bspline}). The outputs are very similar, except for slightly different peak amplitudes, e.g. lowpass filtering yields a higher peak for the right ankle.}
\label{fig:smoothing}
\end{figure}

Finally, small variations between frames cause small high frequency noise. Our experiments showed that a Butterworth low pass filter outperformed other approaches such as convolving with a Gaussian, a Blackman or a Hamming window, or applying a one dimensional Kalman filter. On the other hand it is worth reporting that another method performed very well, which can actually jointly apply imputation and filtering: B-spline curve fitting is a method that fits a polynomial to available data (which can contain gaps). The method is well-suited for the available data, because the player's joints are smoothly transitioning over time yielding a polynomial-like curve. Fig. \ref{fig:bspline} shows that the output (here only the x-coordinate is plotted for simplification) of B-spline fitting is very similar to interpolation plus low-pass filtering (Fig. \ref{fig:lowpass}), with only a few peaks showing different magnitudes (see for example right ankle). A further investigation, i.e. plotting B-spline and low-pass on multiple videos, concluded that both methods seem equally good, as sometimes imputation also interpolates between artifacts where B-spline ignores these artifacts, but on the other hand sometimes B-spline fitting ignores correct peaks in the curve (hand moving up quickly is wrongly regarded as an artifact). Since both methods seem similarly appropriate, it was chosen to employ linear interpolation and low pass filtering for time efficiency reasons.

To summarize, the processing pipeline presented here, starting with raw videos, yields a time series of joint coordinates for each target player separately. In the following, this time series will be referred to as ''joint trajectories'' of one player. In other words, joint trajectories comprise the 2D coordinates of 12 joints (excluding facial key points) of the target person over time (for each frame), which are lowpass-filtered and interpolated already. 

\subsection{Movement classification}

Joint trajectories can be further processed for multiple tasks, including game simulations, but also to classify movements into distinct classes. As mentioned above, we have developed a deep learning approach where the network is trained in a supervised fashion with joint trajectories as the input and the output class being compared to ground-truth labels available from Statcast. Statcast acquires these class labels by manual logging, so any automatic classification is an improvement. The proposed model architecture called MC-CNN is described in detail in section \ref{mccnn} and depicted in Fig. \ref{fig:net_architecture}. The results presented below refer to experiments with smoothed joint trajectories of pitcher and batter of 8245 videos recorded from center field. This camera is used because it is much closer to the players than the side-view camera, leading to a more accurate pose estimation. 

Whilst MC-CNN can be used on other players and even on other sports, here we demonstrate its performance on three important tasks: Inferring the pitching position, the pitch type and the movement of the batter. All accuracies were obtained using ten fold cross validation. We also calculated the average accuracy per class, here called balanced accuracy (BA), because often the number of instances per class varies a lot.

\paragraph{Pitching Position}

Firstly, we want to solve a simple binary classification problem, namely the Pitching Position. In section \ref{motion_analysis}, it was explained what the pitching position refers to and why it is relevant for game analysis. As mentioned in that context, Windup and Stretch differ with respect to speed of the motion and the leg position, so that the classes should be clearly distinguishable from joint trajectories. Accordingly, MC-CNN achieves an accuracy of 97.1\% on average when predicting the pitching position in test data. The balanced accuracy is also 97.0\%, so the approach works equally well for Windup and Stretch. On top of that, a qualitative analysis of errors showed that some of the ''misclassifications'' were actually mislabelled in the dataset.

\paragraph{Pitch type}

Depending on the analysis system, a different number of pitch types is distinguished. In the available metadata there are ten types.\footnote{The metadata contains the following pitch types: Fastball (4-seam), Fastball (2-seam), Fastball (Cut), Fastball (Split-finger), Sinker, Curveball, Slider, Knuckle curve, Knuckleball, Changeup}. While previous work has quite successfully predicted the pitch type from the ball spin and speed, we were interested whether the pitch type is also visible in the general movement, i.e. joint trajectories. It is arguable whether this is possible since not even experts can distinguish between all pitch types without observing the hand grip and the ball. In addition to the problem of different classes corresponding to the same trajectory pattern, some classes have a very high intra-class variability, i.e. the pitch type is executed differently dependent on the player.

\begin{wrapfigure}{R}{10cm}
\begin{center}
\includegraphics[width=10cm]{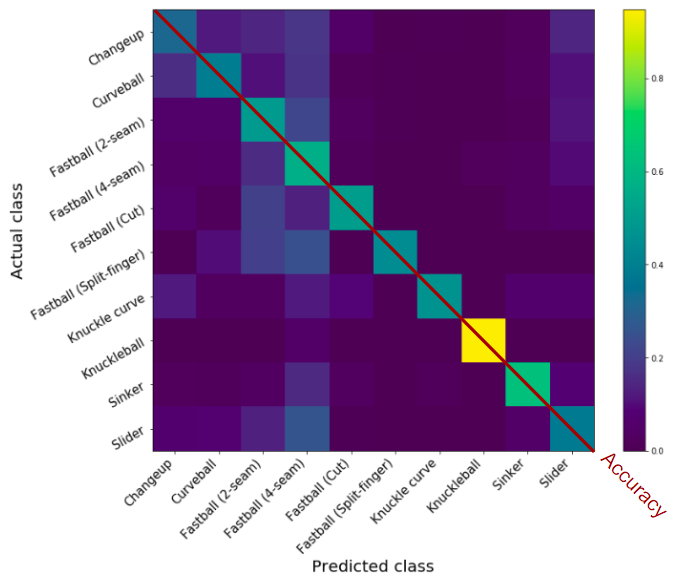}
\caption{The normalized confusion matrix can inform about which pitch types are hardest to distinguish. Rows add up to 1, such that colors indicate the relative confusion with each other. On the diagonal the colour corresponds to the accuracy of predicting this particular pitch type correctly.}
\label{fig:confusion}
\end{center}
\end{wrapfigure}

As expected, our tests could only partly confirm the hypothesis that the pitch type can be classified by the general body motion. Training the network on all ten pitch type, the network achieved 55.8\% accuracy (59.8\%  BA). It easy to see which pitch types are rather similar by analyzing the confusion matrix depicted in Fig. \ref{fig:confusion}. To make sure that the confusion matrix is not due to the inner workings of one particular neural network model, several models were trained and the confusion matrices compared. They appeared to be very consistent, even training with different network architectures (CNN vs RNN) and on distinct subsets of the data. It is thus very likely that the confusion matrix depicted in Fig. \ref{fig:confusion}, which is the average of six models trained separately and tested on different data, is really showing the similarity of the pitch type classes. One can observe for example that Fastball 2-seam and Fastball 4-seam are mixed up sometimes, which makes sense because they mostly differ in speed and motion of the ball. 

For further analysis we varied the training data with respect to the number of classes and variability of players. Interestingly, filtering for Pitching Position, i.e. only taking the videos as input where the pitching position was a Windup (or a Stretch), did not have any effect on accuracy although the position could account for high variability of the joint trajectories. 

Furthermore, we varied the number of players and the number of classes. 
Firstly, in order to find out whether errors are due to differences between players (intra-class differences), we trained the network again taking into account only the five players (starters) for which there were more samples (2519 videos). This dataset contains seven pitch classes, which correspond to the pitch types thrown by these pitchers. Training MC-CNN on this simplified dataset, the accuracy is 65.1\%.  

Of course, the improvement might also be due to the reduced number of classes. Consequently, the next step was to control for inter-class variability, restricting the number of classes. The task was simplified sorting the classes into three superclasses, namely Fastballs, Curveballs and Breaking Balls. The best accuracy, 80.2\%, is achieved in this condition when using only five players on top of that. 

Considering that some pitch types are not distinguishable even for the human observer, this is a very promising result proving the informativeness of joint trajectories. However, as experts can only recognize the pitch type with further input, it might be impossible to reach an accuracy that is large enough for a reliable automatic labeling. Therefore, in future work it will be investigated how additional information can be taken into account. For example, feeding the network with joint trajectories together with the speed of the ball (which is already part of our pipeline) might yield much better results. 

\paragraph{Play outcome}

A third task suited for testing MC-CNN was the play outcome as a variable with three possible assignments: \textit{No swing}, (the batter does not try to hit the ball), a \textit{swing but no run} (the swing resulted in a foul or a strike) and a \textit{run}. Further distinction is not possible solely from the batter's joint trajectories. These three classes are already relevant though to control the cameras and for further processing, e.g. activating ball tracking if the ball was hit. The MC-CNN achieves 97.9\% accuracy (94.7\% BA) on this three class action recognition task. Note also that some misclassifications occur only when localization fails and picks up the wrong person as explained in section \ref{localization_results}. In those videos, the Swing and the Run class are confused.

In general, for each of these three classification problems  we believe some of the errors are due to an unstable pose estimation rather than to the network itself, and hope performance can be improved in an applied system using closer and better quality cameras.

\subsection{Event detection}\label{event_detection}

When evaluating a baseball game, analysts are also interested in important game events to explore the course of a play. For example, these events can be used in systems such as developed in \cite{dietrich2014baseball4d, lage2016statcast, ono2018baseball} to visualize the game timeline. In the Legotracker system, information about the time point of events can also be used to automate camera operation and synchronization. For example, the moment the pitcher starts to move can be seen as the start of a play, meaning that the camera needs to start saving the video, which is later sent to a database storing each play individually. Another possible application is measuring reaction times using the difference between ball release frame and the frame the batter starts running.

Some events are directly visible in joint trajectories, while others can be detected much better incorporating a motion detector. As mentioned in section \ref{overview}, we suggest to let a difference-image approach fill this role accounting for the fast speed of the ball that excludes many other options for motion detection methods such as optical flow. In this section the performance of the proposed framework on event detection is evaluated. 

\subsubsection{Batter's movement}\label{battermovement}

Primarily, the motivation to determine the time point of events in the batter's movement is that it can enable to compute reaction times. Thus, interesting events include the moment the batter puts his foot down to perform a swing, and the frame he starts running, or the "first step". To our knowledge no other system provides such information, so tests had to be conducted on data that we manually labelled ourselves. We took 150 videos in which the batter starts to run and applied a simple gradient thresholding on the joint trajectories to get preliminary results. After manually correcting the videos for which gradient thresholding showed poor performance, a dataset of 150 videos was available for training and testing. In section \ref{battermovement} we explain how we have augmented the dataset, trained a LSTM on the joint trajectories in order to learn the time point of the batter's first step, and further refine the gradient approach to output the moment the batter raises his leg.

During training, test accuracy of predicting the exact frame index of the event was around 25\% (with quite high variance), but in more than 90\% of the data the error margin was less than 3 frames (0.1 seconds). Testing on a separate set of 21 videos, the model achieved a mean square error of 3.43 frames compared to the gradient labels, which is sufficient considering the imprecise definition of a ''first step''. Further, assuming the time of the first step is known, the time frame for the leg lifting can be restricted to a certain range of frames. For 80\% of the test data, it was sufficient to take the maximum of the y coordinate of ankles and knees in this time window as an approximate. In the other 20\% the prediction was slightly late. In further work it might be interesting to explore other approaches, such as training a separate CNN or LSTM on the task of recognizing the leg lifting, or taking a similar approach as for the pitcher's first movement described in the following. In general, other approaches must be explored here, which is a project on its own since inference on the batter's first movement involves labelling data extensively. As the gradient labels are far from perfect themselves, the presented results are rather a proof of concept, demonstrating how such events can be inferred given the body joint trajectories.

\subsubsection{Pitcher's first movement}

\begin{figure}[ht]
\centering
\begin{subfigure}{0.18\textwidth}
    \centering
    \includegraphics[height=9cm]{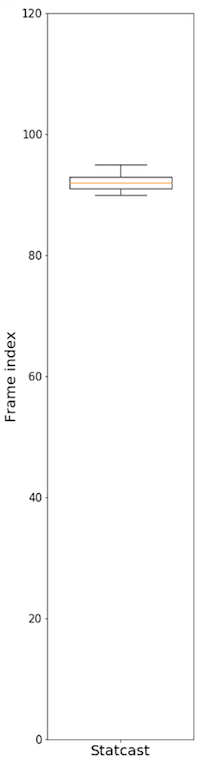}
    \caption{Ball release} 
    \label{fig:release_boxplot}
\end{subfigure}
\hfill
\begin{subfigure}{0.78\textwidth}
    \centering
    \includegraphics[height=9cm]{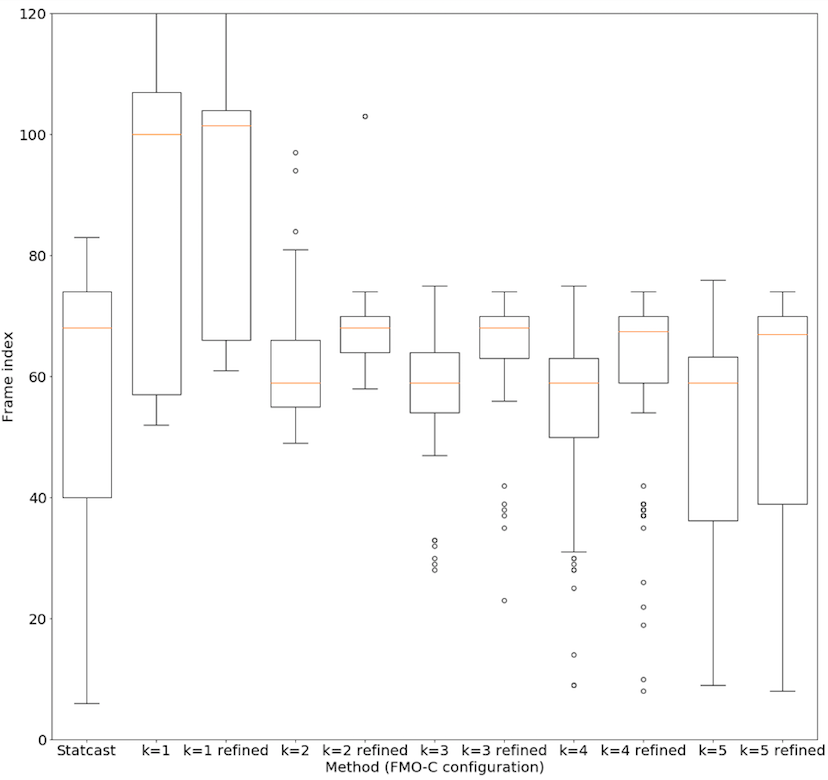}
    \caption{Pitcher's first movement} 
    \label{fig:first_boxplot}
\end{subfigure}
\caption[Self-created figure]{The distribution of the ball release frame index in a sample of videos of 160 frames length is compared to the distribution of the pitcher's first movement frame index. While the release frame is very constant as shown in \subref{fig:release_boxplot}), Statcast labels for the first movement in the left column of \subref{fig:first_boxplot}) vary widely and there are many outliers with unreasonably early motion detection. In comparison, our method FMO-C in different configurations (using every $k$-th frame for different speed sensitivities) shows lower variance, especially using k=2 and k=3. Best performance is achieved by refining the output, taking into account trajectory maxima in a defined range.}
\label{fig:boxplots}
\end{figure}

Labels for the pitcher's first movement are available, but very unreliable. The definition of this event in the metadata is not apparent: In some videos the pitcher has not started to move at all in the ostensible ''frame of first movement'', while in some others his leg is already set back to ground right before pushing the ball forward. Consequently, we decided not to compare our outputs to the ground truth labels directly, but to use another metric: We believe it is more informative to compare the results for 275 center field videos to the ball release frame, thereby indirectly validating the moment of first movement. The release frame is a suitable reference point because the movement of the pitcher is of relatively constant speed, i.e. the time period between first movement and ball release should not vary much. In contrast to the labels for the pitcher's first movement, the available labels for the release frame are very reliable because the videos are roughly aligned at that point. This is apparent in Fig. \ref{fig:release_boxplot} where it is shown that for all videos recording the pitch, the release frame is always around frame 93.

The proposed system recognizes the pitcher's first movement based on motion detection close to the pitcher's leg. As explained in detail in section \ref{first_move}, in our approach a first movement is detected when the difference image method FMO-C finds motion candidates close to the pitcher's leg. Consequently, for this task both joint trajectories and fast moving object candidates (FMO-C) are combined. To run the motion detection method, a hyper parameter $k$ must be set which controls speed sensitivity (cf. \ref{fmoc}). Basically, $k$ defines the frame rate at which difference images are constructed, such that higher $k$ means a lower frame rate and thus larger differences in difference images. Tests were run selecting every $k^{th}$ frame with $k\in [2,5]$, i.e. at a frame rate of 15, 10, 7.5 and 6 fps. In other words, the motion in a time period of 0.07, 0.1, 0.13 and 0.17 seconds respectively is observed. For more details and other hyper parameters see section \ref{first_move}.

In Fig.~\ref{fig:first_boxplot} it is visible how the mean first movement frame index shifts when varying $k$. The higher k, the lower is the artificial frame rate, so inbetween frames there are more changes and thus the difference image indicates more motion. Therefore, $k=1$ is responsive to faster motion and finds the first movement too late, while $k=5$ is too sensitive and leads to many outliers detecting motion in the very beginning of the play. When setting $k=2$ or $k=3$ the approach seems to have the right speed sensitivity to detect a moving leg, because the variance is lower and corresponds to the more realistic assumption that the time between first movement and ball release does not vary much. Highest reliability is achieved ''refining'' the output of $k=2$, $k=3$ or $k=4$, i.e. selecting the highest point of the leg in a range of ten frames around the predicted frame. Setting $k=3$, the approach seems to perform best because for $k=2$ there are some outliers that are too late, which is worse than outliers that are too early when operating the cameras based on this event.

One could of course argue that variance might be due to variance in the data itself. However, plotting the output frame that is predicted with our method (with speed sensitivity $k=3$ refined) for 275 plays, indeed the pose of the pitcher is very consistent. Note that the results comprise both Windups and Stretches, so the method works even if the leg is not raised high (in Stretch position). In contrast, a qualitative analysis of Statcast labels shows large varieties, as sometimes the pitcher stands still, while in other cases the leg is already lifted.

\subsubsection{Ball release frame}\label{pitcherrelease}

In the current state of the framework, the ball release frame is determined by detecting the ball, which is done with the FMO-C method and our GBCV algorithm for ball tracking (cf. \ref{gbcv}). Once the ball is recognized, the time of release can be approximated by its speed and distance to the pitcher.

To evaluate ball tracking, and thus also for concluding about the release frame from the ball appearance, we use videos taken from the side-view camera (Fig.~\ref{fig:side_view}). The reason for this is simply that the ball is hardly visible in front of the heterogeneous audience in the background of the center-field videos, while the side-view videos are filmed from a viewpoint high above the field, so the background is just grass. The set of candidates outputted by FMO-C is thus more reliable, and the ball detection algorithm we developed, GBCV (cf. \ref{gbcv}), is more accurate.

\begin{figure}[ht]
\centering
\includegraphics[height=14cm]{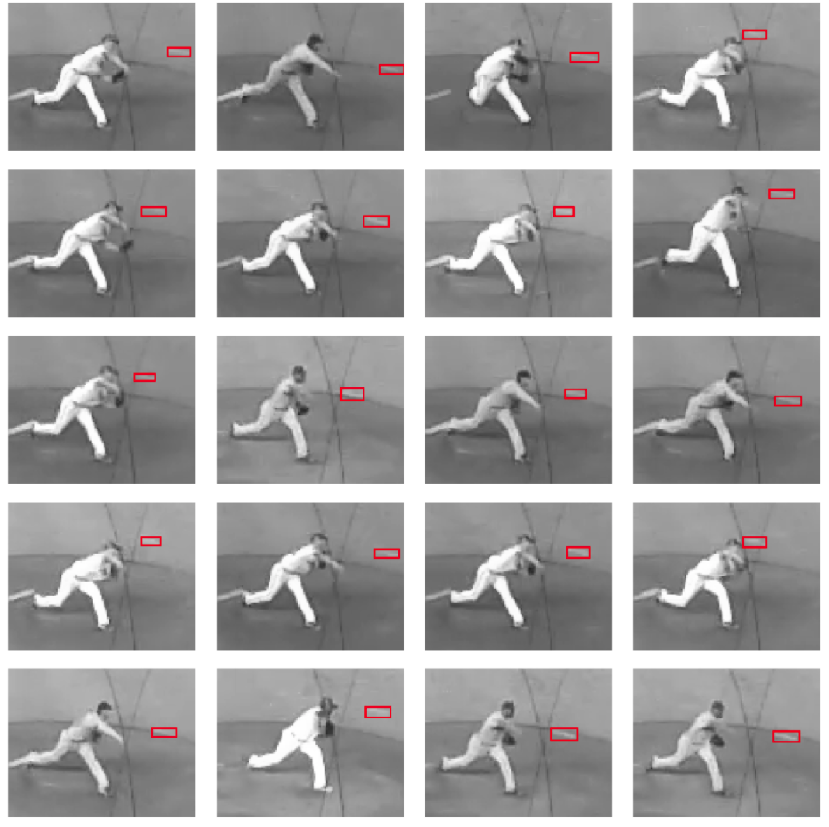}
\caption[Videos of the MLBAM auditing tool (data available through a collaboration with NYU)]{Examples for output frames that are labeled as the release frames with our FMO-C and GBCV approach. One can see that the ball is hardly visible and still the output corresponds to the release frame in all cases, indicating that our ball detection algorithm can distinguish the ball from other moving objects such as the hand.}
\label{fig:all_release}
\end{figure}

Unfortunately, taking side view videos raises the problem that labels are not available in the metadata. Since the side view camera is not synchronized with the center field camera, the available release frame labels are of no use. Therefore, predictions for the release frame by this approach can only be evaluated qualitatively. Looking at the results for hundreds of videos, in approximately 95\% of them the ball was tracked correctly, and once the ball is detected, determining the release frame is straightforward. Considering especially that the ball is hardly visible for the human eye in this video quality, the accuracy is quite remarkable. Examples can be seen in Fig.~\ref{fig:all_release}.

\subsection{Object tracking}\label{objects}

\subsubsection{Ball trajectory and speed} \label{ball}

The task of detecting the ball is challenging, as a baseball is rather small and more importantly very fast, with an average of 92 miles per hour average for Fastballs. On single frames, its appearance is only a blurred grey streak. While tracking the ball is problematic itself because many well-known methods hardly work on this data (cf. \ref{object_detection_background}), the main challenge is detecting the ball in the first place and in particular distinguishing it from other moving parts in the image, most of all the pitcher's arm. 

Due to the speed, the performance of usually well-performing object detection models is very poor. Testing a pre-trained model of Faster R-CNN \cite{Ren} which was trained on the COCO data set that includes baseballs, in less than 10\% videos the ball was recognized at all. As already mentioned, our difference-image approach together with GBCV outperforms other implemented methods by far, detecting the ball in 95\% videos. No ground truth data was available for the ball trajectory, but one can also interpret the following results of speed estimation as a proof of success of the method. 

\subsubsection{Ball speed}

The ball speed was approximated from the 3D trajectory, which in turn was constructed as the intersection of the predicted ball trajectory and a vertical plane from pitcher to batter. Obviously it would be much better to have two or more cameras filming the pitch from different viewpoints and reconstructing the 3D coordinates by combining the output trajectories, but in the available data there were no two cameras with aligned frames. Thus, the missing depth information in the 2D coordinates of the ball must be computed in another way. Since the pitcher throws the ball in a straight line towards the batter, we assume that the ball is somewhere in the vertical plane containing pitcher and batter, and approximate the 3D coordinates as the intersection with this plane.

\begin{figure}[ht]
    \begin{subfigure}{0.5\textwidth}
        \centering
        \includegraphics[width=\textwidth]{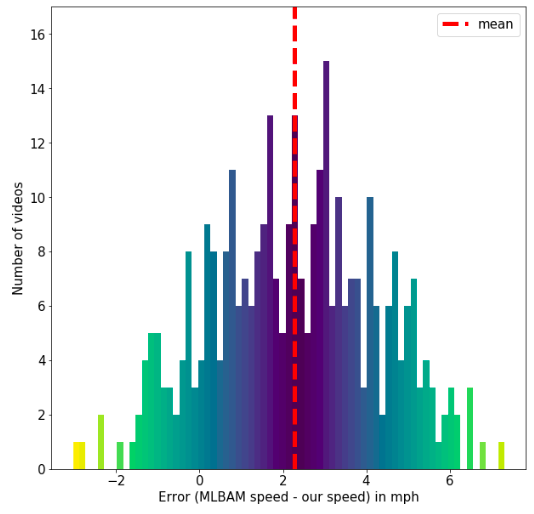}
        \caption{Mean error for camera a}
    \end{subfigure}
    \begin{subfigure}{0.5\textwidth}
        \centering
        \includegraphics[width=\textwidth]{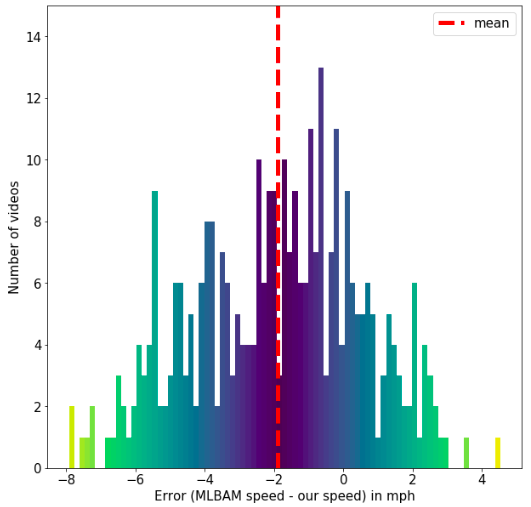}
        \caption{Mean error for camera b}
    \end{subfigure}
\caption[Self-created figure]{Error of speed approximation: The speed is systematically underestimated for camera a, and overestimated for camera b. This might be due to the lack of synchronized cameras, leading to an imprecise approximation of the 3D trajectory.}
\label{fig:ball_histogram}
\end{figure}

The results of speed estimation were evaluated for 331 side-view videos cut to twenty frames, all starting at the ball release. The ball was detected correctly in all videos, with exception of one video where the hand of the pitcher was detected as the ball for a few frames. Fig. \ref{fig:ball_histogram} depicts the error distribution for both available (asynchronous) cameras. Compared to the labels from MLBAM, our calculation had a mean absolute error of 2.53 mph, however systematically underestimating the speed: Subtracting our speed from the ground truth yields an average of 2.27 mph, with a standard deviation is 1.61 mph. A second camera from the other side showed the opposite behaviour, overestimating the speed by 2.5 mph on average. Considering these results, we believe that the error is a consequence of the 3D approximation. It seems that the vertical plane must be shifted towards the second camera. We expect the accuracy will increase once the pitch is shot by synchronized cameras.

\subsubsection{Bat and glove AABB}

Detecting bat and glove is the only task for which we directly used high quality videos, because otherwise the bat is hardly visible. Twenty such videos were available. The glove was detected by the faster R-CNN alone in 62\% of the frames, which is sufficient because the Catcher is not moving much and missing frames can be interpolated.

For bat detection we tested the corresponding module in our framework, a combination of Faster R-CNN and FMO-C (cf. \ref{bat}), taking into account only the frames during the swing (manually selected around 56 frames per videos dependent on swing duration). The faster R-CNN detects the bat in 22.3\% frames which are mostly in the beginning, yielding a suitable starting point for FMO-C. Of the remaining frames, FMO-C detected 57.3\%, yielding an overall detection rate of 66.8\% frames. This is sufficient in all tested videos to approximate 2D tip and base trajectories during the swing. In future work we aim at closing the gaps for example with an ellipse fitting approach.

\section{Integration of units in the Legotracker system}\label{integration_legotracker}

\subsection{Implementation}

The system is implemented in Python, using the OpenCV library for video input processing, and Tensorflow for training NNs\footnote{\url{https://github.com/NinaWie/pitch_type}}. 
For pose estimation we take a model pre-trained on the COCO data set, with a test script in Pytorch.\footnote{\url{https://github.com/tensorboy/pytorch\_Realtime\_Multi-Person\_Pose\_Estimation}}
To detect the baseball bat, we use an implementation of Faster R-CNN available on Github\footnote{\url{https://github.com/rbgirshick/py-faster-rcnn}} in a demo version trained on the COCO data set with Caffe.

\subsection{Integration} \label{integration}

\begin{figure}[ht]
    \centering
    \includegraphics[width=\textwidth]{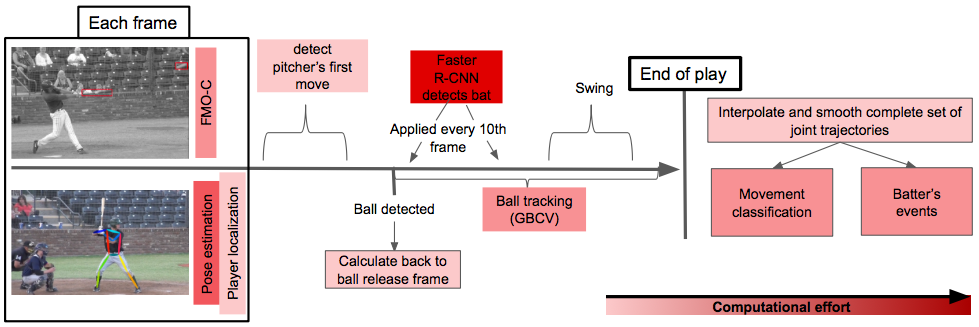}
    \caption[Self-created figure with videos from \url{https://www.youtube.com/channel/UCNFOPbg-VfvJpf7J7x9n4nA} (accessed 29.05.18)]{The sequential processing pipeline of a video is depicted. Pose estimation and FMO-C are applied on each frame, and the output is used for ball and bat detection directly during the video recording. Faster R-CNN and FMO-C outputs together infer the bat trajectory. The complete set of joint trajectories is  filtered in the end and serves as input for the units that cannot be executed online, which is MC-CNN and the finding the batter's events.}
    \label{fig:integration}
\end{figure}

While in the current state of work all presented methods are implemented and tested separately, in the final system they will be integrated and run in parallel. The units can be divided into the ones that directly run while the play is recorded, and the analysis parts that are executed after the course of one play. In the Legotracker setup, each lego block would directly process the video to yield the joint trajectories for their observed target player, but then after the play it would send the data to the database, where further inference can be done (e.g. pitch type classification). For 3D modelling of the pose, several cameras have to observe the same person from different viewpoints. 

In Fig~\ref{fig:integration}, the sequence of processing units and their respective computational effort is visualized. At this point of the project, only pose estimation and the Faster R-CNN are too slow to process a video with 30fps. Summing up the whole processing pipeline, pose estimation is applied on each frame while recording the video, and the target player is directly localized. At the same time, the motion detector FMO-C operates on each triple of consecutive frames (thereby lagging behind two frames). When the processing unit filming the pitcher reports that the play was initialized with the pitcher's first movement, the frame index is reported to the monitoring system, saving the information as the start of the play.

Once the processing unit registers the ball (when the GBCV algorithm outputs that a motion pattern corresponds to a ball trajectory), the ball tracking and speed estimation units are activated. On the other hand, the unit closest to the batter starts preparing for bat tracking. In explanation, the Faster R-CNN model is executed just often enough (inference is too slow to feed each frame) that it yields a start position for bat tracking. FMO-C together with the position of the wrist can then determine the bat trajectory with real time performance. 

The only modules that can not run online, by which we mean directly drawing inference on the current frame or just a few frames later, are the analysis of the batter's events and movement classification in general. MC-CNN as well as the LSTM trained for the batter's first step require a complete set of joint trajectories of one play (a sequence of around 160 frames in the available Statcast data). While neither of both information is crucial for the system to work, unlike for example the pitcher's first movement that initializes further processing steps, of course it would be preferable to provide the information immediately as well. For game analysis purposes, it is definitely sufficient to classify the pitch type after the game and store it in some database, but it might be a motivation for future work that with real time inference, the name of the pitch type could be displayed on the screen straight away, making the information available to the TV audience.

\subsection{Performance}

While on the lowest scale, at a resolution of 368-by-224, pose estimation itself requires on average 0.5 s/frame on a Nvidia Tesla K80 GPU, player localization is insignificant with only 0.0005 s/frame. Interpolation and filtering also takes 0.00085 s/frame, but is not applied in real time anyway because multiple frames are required. Thus, in the final implementation of the Legotracker, the pose estimation module, which is by far the slowest component of pose tracking, should be replaced by better performing models of recent research in this field. Applying the MC-CNN for movement classification only takes 0.0156s without GPU (on a 1.8GHz dual-core processor) for one data point (one play).

Meanwhile, FMO-C operates on grey scale images in parallel to pose estimation. On a Tesla K80 GPU on a video with a resolution of 1080-by-1920 the time effort is 0.12s per frame, including FMO-C, GBCV for ball detection and finding pitcher's first movement. GBCV requires 74\% of the time effort, because a graph of candidates is build and a confidence value must be computed at each iteration. On smaller images of size 540-by-960, the time effort is reduced significantly to 0.033s/frame. This would mean that 30 frames per second can be processed.

\section{Discussion}\label{discussion}

The presented framework describes a possible processing pipeline for a system that captures the baseball game in unprecedented detail. We believe that it was demonstrated that in principle it is possible to acquire all necessary information just from videos, without the need of manual user input. Nevertheless, of course the current version is far from perfect, and work needs to be put into each of the modules in the framework. In general, there are four major challenges that are related to the nature of the data and task: Real-time inference, limited functionality, generalization and 3D approximation. 

Firstly, most processing should run online and provide inference straight away. As outlined in \ref{integration}, most parts in the proposed framework fulfill this criterion. The side effect is that more conventional methods had to be implemented in contrast to deep learning models, because often they are faster and much easier to apply online. For example in bat detection, Faster R-CNN would be too slow (leaving aside that it does not work on blurry images anyways). Employing more deep learning models would definitely make parts of the system more robust and more generalizable. In explanation, pattern recognition with deep learning would enable us to transfer the methods to new videos in an easier fashion, without the need to adapt hyperparameters. The main drawback of the conventional methods implemented here is that they require hyperparameters to be tuned with respect to the type of videos. For example, GBCV requires information about the ball's size on the frame etc. We therefore also considered training a CNN on recognizing the ball in single frames, which would be more robust. Training a CNN might be possible because the ball's appearance is quite constant as a blurry white streak, and once trained, the model would probably be applicable to new videos easier than GBCV. Despite the lack of labeled training data though, the computational effort would most probably be too high to run online on the processing units. In the future the problem might vanish as the computational power of GPUs is still rising, but it seems that currently a game reconstruction framework must rather be a mixture of conventional computer vision and modern Deep Learning.

Secondly, it will be challenging to really achieve sufficient accuracy in all tasks, just with video input. A major step in the pipeline is to infer the player's pose in each frame because all further analysis is much faster on the resulting joint trajectories than on videos. In particular, action recognition (first movement, pitch type) becomes feasible when the data is only a time series data from 12 joints instead of high dimensional video data. However, of course lots of information is lost in this conversion. The experiments on classifying pitching position and play outcome from joint trajectories show that the information is sufficient for some tasks, but for others such as the pitch type the accuracy is too low. For the latter, it is a matter in question whether classification can be achieved from videos of movement in general. Sometimes the pitcher even tries to trick the batter, and pretends to perform a different pitch type, so distinguishing the pitch type is almost impossible even for an expert. Other systems solve the task taking into account the spin rate of the ball, which will hardly be available from video data. As already mentioned, better quality of the videos and closer videos could help a lot, in particular when the motion of the wrist can be observed more closely. 

Third, we do not claim that the modules where Neural Network models were trained could be applied on new video data straight away. The reason is that the dataset available for training was not diverse enough, most importantly comprising videos from just two viewpoints. It is obvious that projecting the video to a set of 2D coordinates makes all inference dependent on the viewpoint of the video. In the current version it is not possible to use a pre-trained model of MC-CNN on new baseball videos to infer to pitch type, as the shape of the joint trajectories changes completely with the viewpoint. This leads to another major point in which the Legotracker system must be extended: A proper system requires 3D coordinates. While two synchronized cameras can be merged to compute a 3D ball trajectory and thereby the ball speed, it is more difficult to do the same on pose estimation data during fast movement with much occlusion. Either this method of camera synchronization, or a model for 3D pose estimation will be crucial to make the Legotracker applicable on a large scale. The long-term goal is thus to develop software such that once the lego blocks and cameras are installed on predefined positions in a new stadium, all components can be executed directly, without requiring further training and tuning of parameters.

\section{Conclusion}

Baseball has already been revolutionized by statistics, but ultimately, stats should be like a third eye for a coach, even analyzing the motion of individuals in detail. The new tracking system called Legotracker is a step towards this goal, using state-of-the-art computer vision techniques to automatically recognize movement, speed and strategy from videos, aiming at full game reconstruction. Our contribution is firstly a framework to incorporate pose estimation in baseball analysis, extracting joint trajectories over time. The results of movement classification on joint trajectories with our proposed model MC-CNN can automate the logging of high-level information of the game, for example denoting strategies such as the pitching position. Furthermore, a fast moving object detector FMO-C and the classifier voting approach GBCV make it possible to reconstruct 3D ball trajectories just from videos. Finally, we achieve a higher reliability in detecting game events than previous systems, combining pose, motion and object detection.

In future work, the presented methods can not only be extended to other players in baseball, but also other sports or action recognition in general. The sports domain is well-suited for action recognition applications, because the number of possible actions is restricted in contrast to real world problems, and a lot of data is available. On the other hand, it should be explored how the presented methods are applicable on completely different tasks. For example, the framework for movement classification might be applicable to video surveillance data. FMO-C on the other hand  could be interesting for autonomous driving where it is extremely important to recognize fast moving objects in real time.

With respect to baseball analysis, most of the proposed methods exceed the accuracy of current systems already or add information that was not provided before. We believe that prospectively it can be improved substantially with videos of higher quality. In addition, the modular approach allows to replace components of the system with state-of-the-art methods. For example, the performance of pose estimation models has improved significantly in recent research work. With further advances in computer vision, even extending the output domain to 3D and thereby also full game reconstruction will finally be possible, such that the experience of sports will be changed significantly.

\section*{Acknowledgements}

Most of all, I want to thank Prof. Silva and Prof. Dietrich for their support. It was a great experience to work on the Legotracker project, and in its progress I always got great advice and insights from all other team members and my supervisors in particular. Furthermore, Estelle Aflalo and Alex Arakelian have contributed a lot with their work on pose estimation and bat recognition - it was a fun and fruitful team work. Last but not least, many thanks to Jannis for proof reading and for your support.

\bibliography{baseball}

\begin{appendices}\label{appendix}

\section{Joint tracking for classification of motion}\label{joint_tracking}

\subsection{Pose estimation}\label{roi}

Instead of inputting the full frame to the pose estimation model, we use a region of interest computed from the person's position in the previous frame. Another possible option would be employing a person detection method on top, but this would require additional computational effort. Instead, we use the last output of the pose estimation to compute the ROI for the next frame. Assuming the target person is already localized in frame $f^t$, the set of 2D coordinates of the defined eighteen body joints (here also enclosing facial keypoints) defines the ROI for the next frame $f^{t+1}$. In detail, the padded axis-aligned bounding box (AABB) enclosing the pose estimation output of the previous frame is taken as the ROI (assuming that the person does not move much between two consecutive frames). Let $X$ be the set of x coordinates of the target player's joints detected in frame $t$, and let $Y$ be the set of y coordinates ($|X|=|Y|=18$). Then the ROI in frame $f^{t+1}$ is defined by the rectangle spanned by the points $p_{1}$ and $p_{2}$, with 
\begin{align}
    p_{1} = (\min(X), \min(Y))^{T} - a,\ \ \ p_{2} = (\max(X), \max(Y))^{T} + a .
\end{align}
$a \in \mathbb R ^2 $ is a vector defining the fatness of the bounding box, which is necessary to account for movement from $f^{t}$ to $f^{t+1}$. Any missing value in $X$ or $Y$ is replaced with the last available coordinate of the respective joint. Otherwise, the bounding box might suddenly shrink and parts of the body would be outside the ROI.

Regarding the first frame, pose estimation is applied on the whole frame, and the position of the target player must be provided beforehand to select one of the detected persons in frame $f^{0}$. The start position of the target person is often known in baseball, for example the pitcher starts in the center of the pitcher's mound.

\subsection{Player localization}\label{localization}

Even in the ROI there might still be other people detected, besides the target player. Thus, in each frame the target player must be found in a list of detected people, given its position in the previous frame. Instead of using the joints directly which might be noisy and contain outliers in some frames, we define the position of a detected person as the bounding box around his (most important / most stable) joints. If $n$ people were detected in frame $f^t$, let $p_{j}^{t}$ be the vector of joint coordinates for person $j$, $j \in [1, n]$. The bounding boxes enclosing each person's joints are then defined as $B(p_{j}^{t})$. We further define the similarity $Sim(p_{j}^{t},\ \hat{p}^{t-1})$ of a detected person $p_{j}^{t}$ to the target person in the previous frame $\hat{p}^{t-1}$ as the intersection over union (IoU) of the bounding boxes around their joints:

\begin{align}
  Sim_{p_{j}^{t},\ \hat{p}^{t-1}} = (B(p_{j}^{t})  \cap B(\hat{p}^{t-1})) / (B(p_{j}^{t})   \cup B(\hat{p}^{t-1})) \;
\end{align}

In other words, the overlap of each detected person with the target person in the previous frame is taken as a measure of similarity.
The new target person $\hat{p}^{t}$ is thus the one with the highest IoU with the previous target person $\hat{p}^{t-1}$, if its IoU overcomes a certain threshold $\theta_{\text{min\_IoU}}$. Otherwise, the joint coordinates are set to missing values (zeros) for this frame.

The main advantage over other approaches is that the threshold $\theta_{\text{min\_IoU}}$ is independent of resolution and camera distance, because the IoU is always between zero and one ($Sim_{p_{j}^{t},\ \hat{p}^{t-1}} \in [0,1]$). In contrast, consider for example the approach of simply selecting the person with minimal absolute pixel-distance to the target in the previous frame, i.e. $\hat{p}^{t-1} = \min_{j} \Vert p_{j}^{t} - \hat{p}^{t-1} \Vert$. Then if a person is not detected at all in one frame, the second closest person would be picked up instead, or a hard threshold must be set, defining the absolute pixel distance that the joints of the target are allowed to move between frames. Our approach is more robust in general, since usually outliers of the joints do not affect the bounding box much, and in addition allows to set a threshold that is generalizing to all kinds of videos, because it is independent of absolute pixel distances.

Furthermore, an upper bound threshold $\theta_{\text{max\_IoU}}$ can be set to account for cases where the pose estimation network mixes up two people, such that we set all joint values of a frame to zero (missing value) if for more than one person holds: $Sim_{p_{j}^{t},\ \hat{p}^{t-1}} > \theta_{\text{max\_IoU}}$. In the frame shown in Fig.~\ref{fig:localize_c} the bounding boxes of both persons overlap a lot with the target (assuming it was detected correctly in the previous frame), so the frame is skipped in the hope that a better distinction is possible in one of the subsequent frames.

In the final version we set $\theta_{\text{min\_IoU}}=0.1$ and $\theta_{\text{max\_IoU}}=0.5$, and only took hips, shoulders, knees and ankles into account to compute the bounding box, because these joints are the most stable ones. The output, namely a time series of joint coordinates for one target player, is imputed with simple linear interpolation and smoothed with low-pass filtering (cf. \ref{filtering_results}), yielding what we call joint trajectories.

\subsection{Movement classification with MC-CNN}\label{mccnn}

\begin{wrapfigure}{R}{0.46\textwidth}
\centering
    \includegraphics[width=0.46\textwidth]{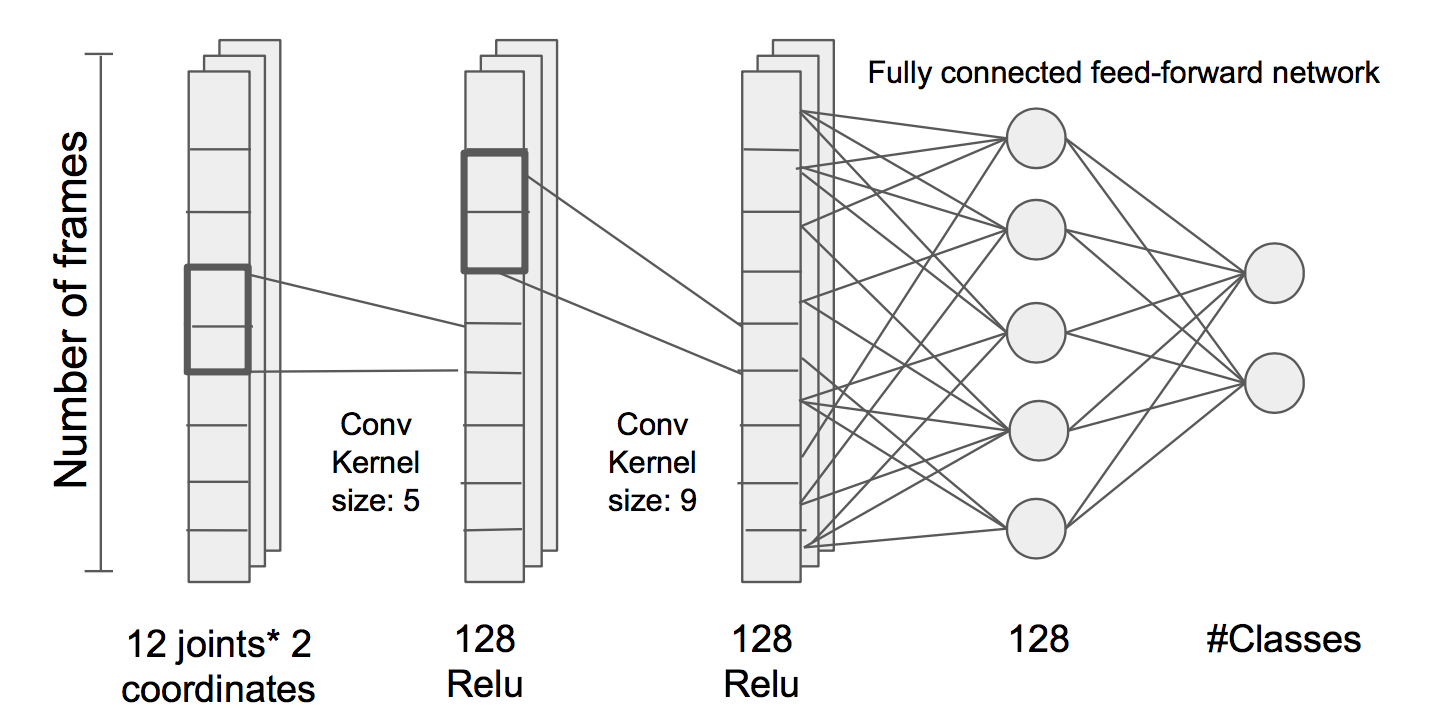}
    \caption{Architecture of MC-CNN}
    \label{fig:net_architecture}
\end{wrapfigure}

We propose a 1D CNN to classify certain motion into discrete classes. The network, in the following called MC-CNN, receives normalized joint trajectories of one player as input and outputs a vector indicating the probability for each possible class. In contrast to inference on video data, processing time is reduced significantly when feeding joint trajectories to a deep learning model. The $x$ and $y$ coordinate of each joint are treated as independent channels, such that the 1D convolutions are applied on 24 channels (12 joints x 2 coordinates), each containing a time-series of one coordinate. 

As depicted in Fig. \ref{fig:net_architecture}, the architecture that performed best consists of two convolutional layers, both with 128 filters and kernel size 5 and 9 respectively, followed by two fully connected layers. The first fully connected layer comprises 128 neurons, while the number of neurons in the second one corresponds to the number of classes, since classes are represented by one-hot encoded vectors. ReLU activation is used for non-linearity in all layers except for the last one where a Softmax function is applied. The network is trained with an Adam Optimizer minimizing a cross entropy loss function with a learning rate of 0.0005. The network was trained for 2000 epochs, although convergence seems to be reached after around 200 epochs.

Furthermore, we balance the batches (of size 40) such that the number of examples per class in a batch is constant. For example, for pitch type classification with 10 pitch type classes, this corresponds to 4 samples per class in each batch. Balancing leads to a higher accuracy as the network does not overfit as much on the classes that appear most often in the data. Last, the time series data is normalized independently for each channel, such that the time series values of each coordinate of each joint have mean zero and unit variance.

\section{Fast Moving Object Candidate detection (FMO-C)}\label{fmoc}

Inspired by the work of \citet{Rozumnyi2017}, we developed a method that detects objects of high velocity, called FMO-C. Similarly to \citet{Rozumnyi2017}, FMO-C operates on three consecutive frames, thresholding their difference images and searching for connected components. Our variation from the original approach is 1) we allow for different speed sensitivities taking every k-th frame into account, and 2) we compensate for jitter.

At each time point, the input is a set of three frames in grey scale. However, these frames are not necessarily consecutive. Let $f^{1}, ..., f^{n}$ be all frames of a play. Firstly, for each frame $f^{t}$, the three possible difference images $d$ between $f^{t-k}, f^{t}$ and $f^{t+k}\ (k<t \leq n-k)$ are computed as
\begin{align}
d^{i,\ j} =  \theta \lvert (f^{i} - f^{j})\rvert\ .
\end{align}

Thereby, $k \in \mathbb N,\ k>0$ is used to control the speed sensitivity, because selecting only every $k^{th}$ frame affects the difference images: the higher $k$, the smaller is the artificial frame rate, and the larger is the difference between frames. Thus, the higher $k$, the more motion is detected. $k$ can then be set with respect to the task. For example, $k$ should be smaller for ball detection than for the pitcher's leg's motion, because for small k, only very fast motion is detected, and the ball with its high velocity is recognized easily. 

\begin{figure}[ht]
    \centering
    \begin{subfigure}{0.32\textwidth}
    \includegraphics[width=\textwidth]{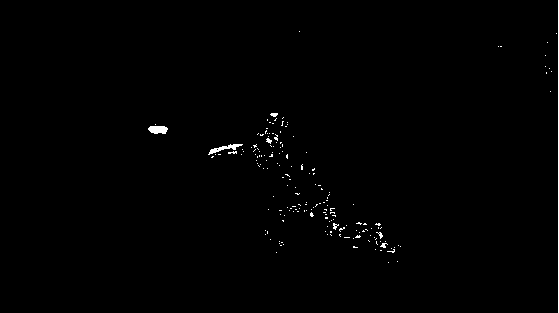}
    \caption{Thresholded difference}
    \label{fig:fmo1}
    \end{subfigure}
    \hfill
    \begin{subfigure}{0.32\textwidth}
    \includegraphics[width=\textwidth]{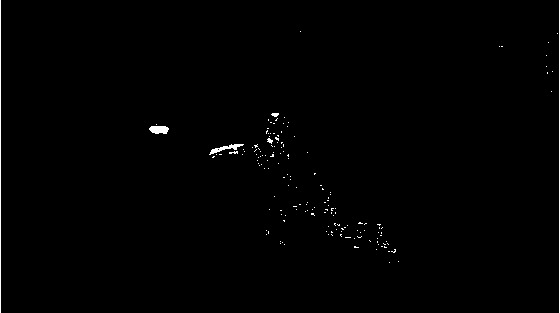}
    \caption{Jitter removed}
    \label{fig:fmo2}
    \end{subfigure}
    \hfill
    \begin{subfigure}{0.32\textwidth}
    \includegraphics[width=\textwidth]{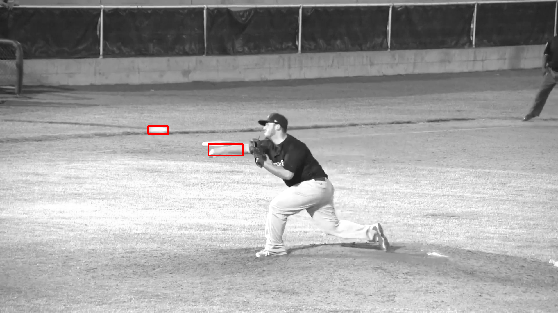}
    \caption{Connected components}
    \label{fig:fmo3}
    \end{subfigure}
\caption{FMO-C: Firstly, a simple difference image is thresholded (\subref{fig:fmo1}). To account for jitter, accumulated previous movement is removed (\subref{fig:fmo2}). In the end, only connected components of a certain minimum area are selected as candidates, which are marked red in (\subref{fig:fmo3}). This leads to a set of candidates of fast moving objects for each frame.} \label{fig:fmo_detecton}
\end{figure}

The reason for taking three pictures into account is that a difference image between two frames picks up both the previous position and the new position of a moving object. With three images, the previous location can be excluded by logically combining the difference images. Formally,
\begin{align}
    \mathbb{D}^t = d^{t-k,\ t} \cap d^{t,\ t+k} \cap \neg d^{t-k,\ t+k}.
\end{align} 
The result is just one difference image $\mathbb{D}^t$, containing only the appearance of motion in the target frame $t$ as shown in Fig.~\ref{fig:fmo1}.

Furthermore, the method should be robust to slight motions of the camera. Although the cameras are fixed, vibrations of the stadium might cause noise in the difference images. Here we can make use of the fact that an unstable camera is moving randomly around one location, but the objects we are interested in move one-directional for m consecutive frames. Thus all points in $\mathbb{D}$ that were also detected in one of the last $m$ frames can be excluded:  
\begin{align}
\mathbb{F}^i = \mathbb{D}^i -  \bigcup_{n\in [i-m, ..., i-1]} \mathbb{D}^n
\end{align}

Nevertheless there are still many artifacts left, especially many single pixels. Therefore, as in \cite{Rozumnyi2017}, a threshold $\theta_{\text{conn}}$ is introduced defining the minimum area that a moving object must cover. $\theta_{\text{conn}}$ can be set dependent on image resolution, camera distance and optical zoom. In our implementation we apply the OpenCV function \texttt{connectedComponentsWithStats} on $\mathbb{F}^i$ to compute the AABB and area of each moving object, and filter out all components with less pixels than $\theta_{conn}$ (see Fig. \ref{fig:fmo3}). The output is a set of patches (''motion candidates'') as in Fig.~\ref{fig:fmo3}, each covering a certain minimum area. 

\section{Event detection}

\subsection{Batter's first step}\label{batter_first_methods}

Given the video of a play, our goal is to output the index of the frame when the batter takes the first step, given the set of joint trajectories of the batter. The notion of a first step is not clearly defined, as sometimes it is hard to distinguish between the end of the swing and the first step. Because of this, FMO-C is not applicable: The motion candidates during the swing do not differ from the ones appearing when the batter starts to run towards 1st base. Furthermore, training an ANN on images or joint trajectories is not possible straight away, because no ground truth labels are available so far. Thus, rather basic methods are applied on the joint trajectories in the first place, namely gradient-based methods. Thresholding the gradient of the $x$ coordinates is most informative because the batter moves to one side when starting to run. For most of the videos, reasonable results were achieved: Manually observing the outputs, the result seemed to deviate from the ground truth moment only by around four frames. However, for the ones that did not overcome the threshold, the results were far outliers or no result at all. A slight improvement was achieved when iteratively lowering the threshold until a frame is found, but the results are still highly dependent on the video material. To avoid such a hard threshold, and to make the method generalize better, we used the outputs of the gradient approach as training data for an ANN. Firstly, we manually labeled the ones that were mislabeled by the gradient approach. For the input set we only selected the frames plausible as a first step, assuming that the ball release frame is known. In explanation, we only input a window of 40 frames to the LSTM, starting 10 frames after the (estimated) release frame $f^{r}$. We chose this window of $[r+10, r+50]$ because we observed thar the first step occurs on average 30 frames after ball release, with a variance of around 5 frames. Spanning the window by twenty frames in each direction accounts sufficiently for errors in measurement of the release frame, or for outliers of the first step frame index.

This yields a dataset of joint trajectories of 40 frames length each, annotated with the first-step-frame-index. Furthermore, we artificially augmented the data by shifting the frames in time: the window of 40 frames was randomly placed $k$ times for each data point, such that the first step frame was uniformly distributed between 0 and 40. Finally, we flipped the $x$ coordinate for each data point (doubling the amount of data), such that a left to right movement is turned into a right to left movement. On the resulting dataset, best performance is achieved by a LSTM of four cells with 128 hidden units each, followed by one fully connected layer. The output is a number $y \in [0, 1]$ that can be transformed in the following way to yield the frame index $s$ of the frame depicting the first step: 
\begin{align}
s = 40 y + r + 10\ .
\end{align}

\subsection{Raise of the batter's leg}

The other relevant part of the batter's movement is the moment he lifts his front leg and puts it back to the ground. The lifting of the leg initiates the swing and thus usually occurs slightly before ball release. Similarly to the first step, the time frame for this event is very restricted when the time points of other events are known. Therefore it is sufficient to determine the leg-raise-frame in a window of frames, for example from twenty frame before ball release $r$ to ten frames before the first step $s$. In this period we define the leg-raise frame simply to be the frame where the batter's leg (ankle and knee joints) are highest:
\begin{align}
l = \argmin{t \in [r-20,\ s-10]}y^{t},
\end{align}
where $y^{t}$ is the mean $y$ coordinate of both ankles and knees at frame $f^t$. Minimum instead of maximum is taken because $y$ is zero at points on top of the frame. In addition to finding frame index $l$ where the leg is highest, it is relevant for analysis purposes to infer an event slightly later, when the foot is put back to the ground. In order to find this moment, firstly a reference point for the foot position on the ground is required. The baseline position $m$ of the leg, i.e. the average position before lifting it, can be computed as:

\begin{align}
m = \dfrac{1}{l-10} {\sum\limits_{n=0}^{l-10} }y^{n}
\end{align}

The frame $h$ when the foot is put back to ground is then the frame out of all $r$ frames following $f^l$, where the leg position is closest to $m$:

\begin{align}
    g = \argmin{i \in [l,\ l+r]}  \lvert y^{t}-m \rvert
\end{align}

The range r is dependent on the frame rate, but usually around 10-20 frames should be sufficient because it does not take longer to set the foot back.

\subsection{Pitcher's first movement}\label{first_move}

The moment the pitcher starts to move can be seen as the start of a play, and it is often taken as the reference time for the computation of several statistics. In order to capture the full movement of the pitcher, we define the ``pitcher's first movement'' as the moment the leg is raised. To find the corresponding frame, we combine candidates of FMO-C with pose estimation. The idea is that when the pitcher starts to move, a motion detector should find candidates at the pitcher's leg in several consecutive frames. 

\begin{figure}[t]
    \centering
    \includegraphics[width=\textwidth]{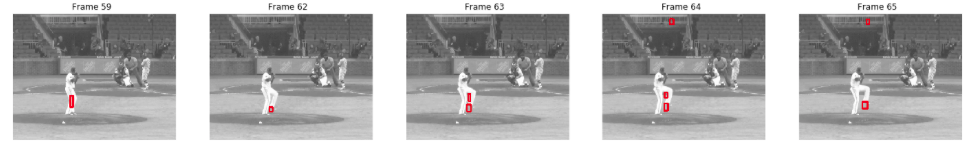}
    \caption{An example sequence of frames with motion candidates close to leg is shown. In frame 56, the pitcher moves slightly but the frame is isolated, while frame 62 would be labeled as the pitcher's first movement in case all requirements are fulfilled.}
    \label{fig:first_move_candidates}
\end{figure}

We use the set of left and right ankle coordinates $a_{l}, a_{r} \in A^{i} $ and knees $k_{l}, k_{r} \in K^{i} $ at each frame $f^{i}$, evaluating their distance to the motion candidates $c \in C^{i}$ detected in $f^{i}$, where $\forall (v \in A \cup K \cup C) : \vec{v} \in \mathbb R ^2 $. For a single frame, we say that it is likely to be part of the first-movement frame sequence if a motion candidate $c$ is close to the ankles or knees, whereby ''closeness'' is defined as a fraction of the distance between ankles and frames. Formally, the condition can be written as

\begin{equation}
    \exists \vec{u} \in \{a_{l}, a_{r}, k_{l}, k_{r}\} \wedge \exists \vec{c} \in C^{i} : \| \vec{u} - \vec{c} \| < \dfrac{1}{2} b \sum_{j \in \{l,r]\}} \| a_{j} - k_{j} \|.
    \label{equation_pitcherfirst}
\end{equation}

The right side of the condition in \autoref{equation_pitcherfirst} defines the required closeness to the ankles/knees. As mentioned above, the radius itself is defined by the distance between ankles and knees in order to construct a threshold that is independent of video resolution and distance from the player. The radius is factorized by a parameter $b$ that can be set based on the video data quality and the accuracy of the pose estimation.  Here, for our low quality videos recorded from larger distance, pose estimation is quite inaccurate, so we set $b=1$ such that the radius is simply the mean distance between knees and ankles.

The first-movement frame is then the beginning of a set of frames for which \autoref{equation_pitcherfirst} holds, where the set of frames is restricted in two ways: The sequence containing this set must comprise at least $\theta_{min\_length}$ frames, and the first and the last frame (where the condition is fulfilled) must be less than $\theta_{max\_apart}$ apart. The first threshold ensures that a minor leg motion long before the actual first movement is not picked up, while  $\theta_{max\_apart}$ makes sure that the real first movement is detected even if there are gaps (no detection) of less than $\theta_{max\_apart} - \theta_{min\_length}$ frames inbetween. In the example depicted in Fig.~\ref{fig:first_move_candidates} one can see that in frame 59 some motion is detected already, but only from 62 onwards the movement really starts. In our experiments, we set $\theta_{\text{min\_length}}$ to 5 and $\theta_{\text{max\_apart}}$ to 10, so since $65-59<10$, the criteria are fulfilled and the method would output frame 59 as the first-movement frame.

In the final step we refine this output to achieve an even more constant definition of the first movement. For this purpose, the curvature of the joint trajectory can be taken into account. In detail, ''refining'' refers to taking the output from the algorithm explained above, and selecting a more stable point from a window of frames around the previously predicted frame. A simple approach is selecting the highest position of the leg (mean of ankles and knees) in a certain range $p$ around the previously predicted frame index $n$, formally 
\begin{align}
    h =  \frac{1}{4} \argmin{t \in [n-p,\ n+p]}\  \sum_{v \in A^{t} \cup K^{t}} v_2\ \ \ \ \ \ (v_2\ :=\ \text{y coordinate of joint v}).
\label{equationRange}
\end{align}
Consequently, $f^{h}$ is the moment the leg is highest, which is a sharper definition of the first movement and quantifies it more precisely.

\section{Object detection}\label{object_detecion_methods}

\subsection{Ball detection}\label{gbcv}

The main challenge in ball detection is distinguishing it from other moving objects with similar appearance. Most of all, the hand of the pitcher becomes almost as blurry and greyish when releasing the ball. As a possible solution, we propose a graph based voting of weak classifiers considering several features of the ball trajectory. 

The algorithm explained in the following operates on the output of FMO-C (which is a set of motion candidates per frame). Firstly, each candidate is represented by a node in a directed acyclic graph. Specifically, the graph is a tree where each level in the tree corresponds to a frame. Let $n^{t}_{j},\ j\in[1..n]$ be the j-th motion candidate detected in frame $f^t$. Then $n^{t}_{j}$ is a child of some node $n^{t-1}_{k}$ of the level above (the frame before) iff the corresponding candidates are more than $\theta_{dist}$ pixels apart. This is based on the assumption that the ball travels with a certain minimum speed. Many candidates can be excluded straight away if their speed is too low, and thus computational effort can be reduced. The threshold $\theta_{dist}$ can be deduced from the minimum speed of the ball in a pitch, the frame rate, the distance from the camera and the resolution. For the experiments here it was set to 10 pixels. Consequently, a node has no children if there was no candidate detected in the next frame or all candidates were too close.

In the resulting tree, each traverse of length $\geq 3$ is a possible ball trajectory. To distinguish the ball from the large set of other paths in the graph, we define a confidence value C as a combination of several attributes of the ball trajectory. A possible choice for such attributes are slopes and distances between consecutive motion candidates, because approximately they stay constant only if it is the ball, assuming a high frame rate and a relatively high speed. In other words, we take three consecutive frames, compute slope and distances between the candidates of the first and the second one, and the same for the second and third one, and measure how similar they are. Formally, let $s(i,j)$ be the slope of two connected nodes (candidates) $i$ and $j$, and $d(i,j)$ the distances between each pair. Then a triple of three connected nodes, i.e. a node $n^{t-2}$ with a child candidate detected in $f^{t-1}$ and a grandchild in $f^{t}$, is classified as a ball if the confidence $C$ is sufficiently high. $C$ combines and weights the defined attributes, and only if $C$ exceeds a threshold $\theta_{\text{confidence}}$, the triple of motion candidates is recognized as a ball. 

In our implementation, the confidence value C is defined as: 

\begin{equation}
\begin{split}
    C([n^{t-2},\ n^{t-1},\ n^{t}]) =\  & a_{1}\ S_{\text{slopes}}(s(n^{t-2},\ n^{t-1}),\ s(n^{t-1},\ n^{t}))\ +\\
    & a_{2}\ S_{\text{distances}}(d(n^{t-2},\ n^{t-1}),\ d(n^{t-1},\ n^{t})).
\end{split}
\end{equation}

$S_{\text{slopes}}$ is a measure for the similarity of two slopes, and $S_{\text{distances}}$ a measure to compare two distances. In addition, the attributes can be weighted with a vector $\vec{a}$. 

To construct a similarity measures $S$, some requirements should be fulfilled: Ideally, the two similarity measures should be comparable, such that they take on the same range of values. Secondly, the confidence value and thus the similarity measures should be independent of the data properties (e.g. resolution and distance of the camera). To account for both, we define the slope $s$ as a complex normalized vector, because if the distance in x and y direction were simply divided, the slope for a vector $\vec{v}$ would be the same as the slope for the vector in the opposite direction $\vec{-v}$. This case must be considered because FMO-C might for example first detect motion at the leg, then the arm in the next frame and then again the leg. Both the distances and the slopes between this triple of nodes in the graph would be the same, but defining the slope as a complex vector, their value is different. So the difference between two slopes is the distance of two normalized complex vectors. The similarity is then disproportional to the distance of the respective complex vectors of each slope:

\begin{align}
    S_{\text{slopes}}(s_{1}, s_{2}) = 1\ -\ (\frac{1}{2}\ \Vert s_{1}\ -\ s_{2}\Vert)
\end{align}

The formula for $S_{\text{slopes}}$ is thereby constructed to yield one for equal slopes, and zero for vectors in the opposite direction.

A similarly standardized value should define the similarity of distances. Furthermore, we want to ensure that only the relative distance is considered (independent of pixel values), so instead the ratio of distances is considered:

 \begin{align}
    S_{\text{distances}}(d_{1}, d_{2}) = \min (\frac{d_{1}}{d_{2}},\frac{d_{2}}{d_{1}})
\end{align}

\begin{figure}[ht]
\begin{subfigure}{0.5\textwidth}
    \centering
    \includegraphics[width=\textwidth]{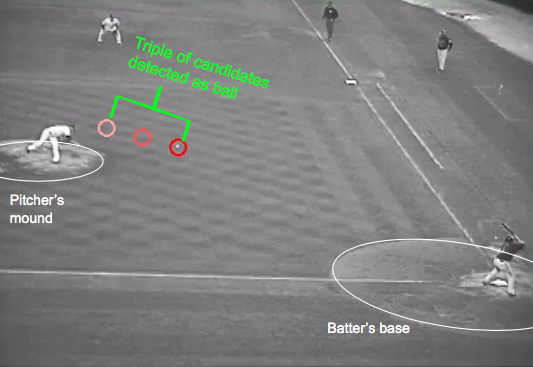}
    \caption{\label{fig:trajectory1}}
\end{subfigure}
\hfill
\begin{subfigure}{0.5\textwidth}
    \centering
    \includegraphics[width=\textwidth]{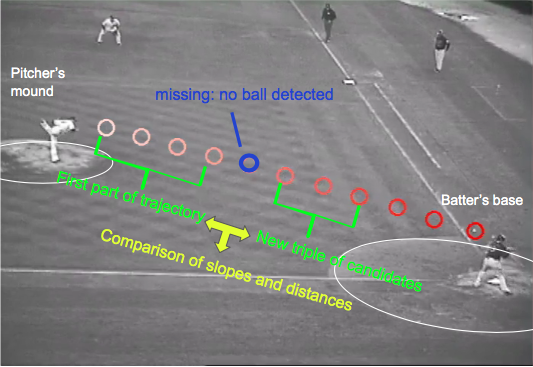}
    \caption{\label{fig:trajectory2}}
\end{subfigure}
\caption[Self-created figure, with videos from the public MLBAM database at \url{http://ze-video.mlb.com/video/mlbam/2016/10/01/umpeval/video/449253/} (accessed 13.05.18)]{In (\subref{fig:trajectory1}) the ball is detected, i.e. the slopes and distances of the three ball candidates were sufficiently similar. Then in one frame the ball is not detected, but from the next frame onwards GBCV registers a new triple corresponding to a ball trajectory. By comparing slopes and distances of the new triple to the previous trajectory, the trajectories can be merged. The ball is tracked until it reaches the batter (\subref{fig:trajectory2}).}.
\label{fig:ball_trajectory}
\end{figure}

The minimum of both ratios of distances again yields an output between 0 and 1, and also makes $S_{\text{distances}}$ symmetric. To sum up, the confidence value is defined in a way that different measures of similarity can be combined flexibly and the ranges of output values are similar. Depending on the video material, other attributes can be incorporated, for example the area of the bounding box enclosing a candidate or the average color of the image patch of a candidate that is supposedly white or greyish. Also, similarly to the threshold for the pitcher's first movement, a minimum sequence length can be set to avoid false positives, such that three nodes are not sufficient. An example of a triple of FMO candidates that is recognized as a ball by $C$ is is shown in Fig.~\ref{fig:trajectory1}.

Gaps can occur if the ball is not detected by FMO-C in one or more frames. If later again three ball candidates are found, average slopes and distances of the two separate trajectories can be compared in the same fashion as before, and merged if $C$ is sufficiently small. This is illustrated in Fig.~\ref{fig:trajectory2}.

\subsection{Bat and glove AABB}\label{bat}

A successful approach locating objects in images is called Faster R-CNN \cite{Ren}, and it can be trained on the COCO data set to recognize bat and glove. Testing a pre-trained model on our videos, we observed glove detection is sufficient for our purposes, but the bat was often not detected in the crucial moment of the swing itself. This might be due to the fact that images of blurred bats are hardly represented in the databases. We therefore propose a combination of Faster R-CNN and the FMO-C approach. Once the bat starts to move and is not detected anymore by Faster R-CNN, FMO-C takes over. Formally, from the set of candidates $C^{t}$ in frame $f^t$, the baseball bat can be found by simply taking the detection with the shortest Euclidean distance to the previous bat detection $\beta^{t-1}$. To avoid unreliable candidates that are too far away from the previous detection, a threshold $\theta_{\text{max\_dist}}$ is used:

\begin{align}
    \beta^{t} = 
\begin{cases}
    \argmin{c_{k} \in C^{t}}\ \|(c_{k}-\beta^{t-1})\|,& \text{if } \underset{c_{k} \in C^{t}}\min\ \|(c_{k}-\beta^{t-1})\|\ <\ \theta_{max\_dist}\\
    \text{missing},              & \text{otherwise}
\end{cases}
\end{align}

\begin{figure}[ht]
\begin{subfigure}{0.5\textwidth}
    \centering
    \includegraphics[height = 3cm]{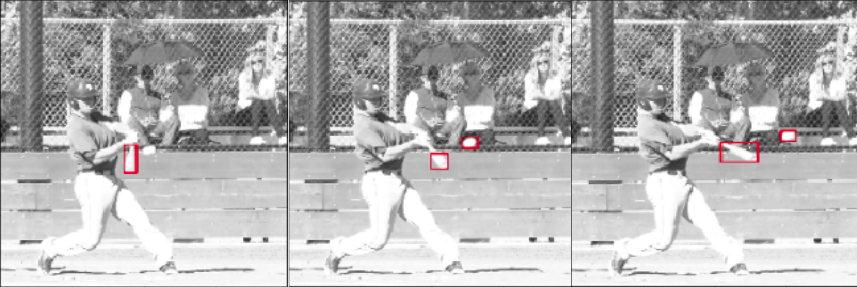}
    \caption{The FMO-C algorithm outputs a bounding box for each motion candidate.} \label{fig:fmo_bat}
\end{subfigure}
\hspace*{0.1\textwidth}
\begin{subfigure}{0.4\textwidth}
    \centering
    \includegraphics[height = 3cm]{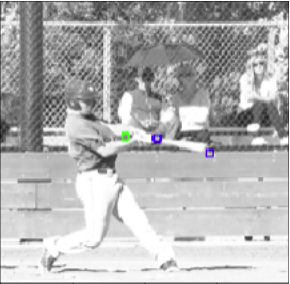}
    \caption{The wrist position coloured green is used to differentiate between tip and base of the bat.} \label{fig:wrist_tip_bat}
\end{subfigure}
\caption{Bat detection during the swing: First, an object detection method is employed to detect the bat as a reference point. Then, FMO detection is applied during the swing (see \subref{fig:fmo_bat}), and the candidate closest to the previous bat detection is selected. Finally the position of the wrist enables us to distinguish tip and base of the bat (\subref{fig:wrist_tip_bat}).}\label{fig:swing}
\end{figure}

So if $\beta^{t-1}$ is given (detected either by the Faster R-CNN or by FMO-C), each motion candidate in frame $t$ is compared to $\beta^{t-1}$ and set as the new bat position if it is sufficiently close. In Fig.~\ref{fig:fmo_bat}, outputs of FMO-C detection are shown. Both Faster R-CNN and FMO-C yield the AABB around the bat. 

However, the orientation of the bat in this bounding box is necessary for a more detailed description of the bat trajectory and for speed estimation. In explanation, the aim is to recover the positions of tip and base of the bat separately. This can be achieved taking into account the wrist coordinates available from pose estimation, assuming that the corner of the AABB which is closest to the wrists is the base of the bat, and the opposite diagonal corner is the tip. In Fig.~\ref{fig:wrist_tip_bat} the wrist is coloured green, leading to the location of tip and base of the bat (blue).
 
Finally, with this combination of the Faster R-CNN, FMO detection and pose estimation, the 2D trajectory for tip and base of the bat can be estimated for the full length of the swing. 
Further research needs to be done to turn this into a 3D trajectory to compute speed.

\end{appendices}

\end{document}